\documentclass{article}[12pt]

\usepackage[utf8]{inputenc} 
\usepackage[T1]{fontenc}    
\usepackage{url}            
\usepackage{booktabs}       
\usepackage{amsfonts}       
\usepackage{nicefrac}       
\usepackage{microtype}      
\usepackage{amsmath,amsthm}
\usepackage[numbers]{natbib}
\usepackage{authblk} 

\usepackage[hidelinks]{hyperref}
\usepackage[all]{hypcap}
\hypersetup{citecolor=blue}
\hypersetup{linkcolor=black}
\hypersetup{urlcolor=MidnightBlue}

\usepackage[nameinlink]{cleveref}
\usepackage[algoruled]{algorithm2e}
\usepackage{epsfig}
\usepackage{subcaption}
\usepackage{dsfont}
\usepackage{xr}
\crefname{algocf}{Alg.}{Algs.}
\Crefname{algocf}{Algorithm}{Algorithms}
\crefname{appsec}{app.}{apps.}
\Crefname{appsec}{A.}{A.}

\newtheorem{theorem}{Theorem}
\newtheorem{lemma}{Lemma}

\newcommand{\N}{\mbox{N}}

\newcommand{\R}{\mathbb{R}}
\newcommand{\vnorm}[1]{\left|\left|#1\right|\right|}

\title{A Bayesian Model of Dose-Response for Cancer Drug Studies}
\date{}

\author[1]{\small Wesley Tansey\thanks{\texttt{tanseyw@mskcc.org} (corresponding author)}}
\author[2]{\small Christopher Tosh}
\author[2,3,4]{\small David M.~Blei}

\affil[1]{\footnotesize Department of Epidemiology \& Biostatistics, Memorial Sloan Kettering Cancer Center, New York, NY, USA}
\affil[2]{\footnotesize Data Science Institute, Columbia University, New York, NY, USA}
\affil[3]{\footnotesize Department of Statistics, Columbia University, New York, NY, USA}
\affil[4]{\footnotesize Department of Computer Science, Columbia University, New York, NY, USA}

\begin{document}

\maketitle

\begin{abstract}
Exploratory cancer drug studies test multiple tumor cell lines against multiple candidate drugs. The goal in each paired (cell line, drug) experiment is to map out the dose-response curve of the cell line as the dose level of the drug increases. We propose Bayesian Tensor Filtering (BTF), a hierarchical Bayesian model for dose-response modeling in multi-sample, multi-treatment cancer drug studies. BTF uses low-dimensional embeddings to share statistical strength between similar drugs and similar cell lines. Structured shrinkage priors in BTF encourage smoothness in the dose-response curves while remaining adaptive to sharp jumps when the data call for it. We focus on a pair of cancer drug studies exhibiting a particular pathology in their experimental design, leading us to a non-conjugate monotone mixture-of-Gammas likelihood. To perform posterior inference, we develop a variant of the elliptical slice sampling algorithm for sampling from linearly-constrained multivariate normal priors with non-conjugate likelihoods. In benchmarks, BTF outperforms state-of-the-art methods for covariance regression and dynamic Poisson matrix factorization. On the two cancer drug studies, BTF outperforms the current standard approach in biology and reveals potential new biomarkers of drug sensitivity in cancer. Code is available at \url{https://github.com/tansey/functionalmf}.
\end{abstract}

\section{Introduction}
\label{sec:intro}

To search for new therapeutics, biologists carry out exploratory
studies of drugs.  They test multiple drugs, at different doses,
against multiple biological samples.
The goal is to trace the dose-response curves, and to understand the
efficacy of each drug.

This article concerns dose-response modeling in exploratory drug studies.
In particular, we are concerned with studies where the experimental design makes it difficult to perform statistical inference on the resulting data.
We consider two such studies, both involving anti-cancer drugs being tested \textit{in vitro} on
models of human tumors known as organoids \citep{drost:clevers:2018:organoids}. A dose-response curve in the
studies represents the expected cell survival rate (response) for a specific organoid
as a function of the concentration (dose) of an anti-cancer drug. The studies differ primarily in their size.
The first study
is a small-scale pilot study conducted internally at Columbia University Medical Center
with $35$ drugs and $28$ organoids;
the second is a large-scale, ``landscape'' study conducted at Samsung Medical Center
with $67$ drugs and $284$ organoids \citep{lee:etal:2018:gbm-organoids-drug-response}.
The experiments in each study are costly; each one can
take weeks or months to conduct in the lab. Consequently, the exhaustive
set of all (organoid, drug) combinations is not available.
This leaves missing data, dose-response curves for which no data is available,
that must be imputed.

\begin{figure}[t]
\centering
\begin{subfigure}{0.3\linewidth}\includegraphics[width=0.97\textwidth]{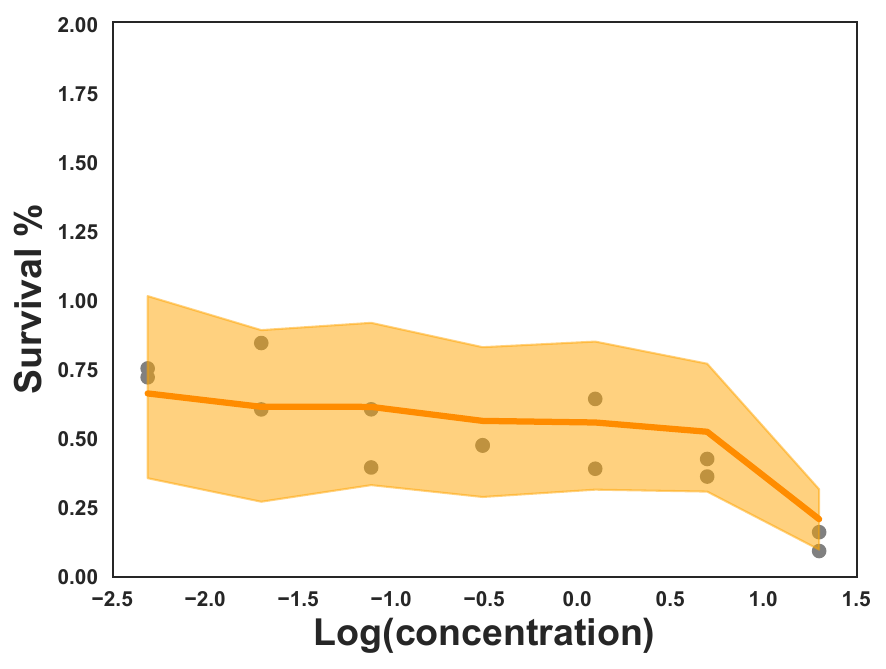}\caption{Organoid 1, Drug 1}\end{subfigure}
\begin{subfigure}{0.3\linewidth}\includegraphics[width=0.97\textwidth]{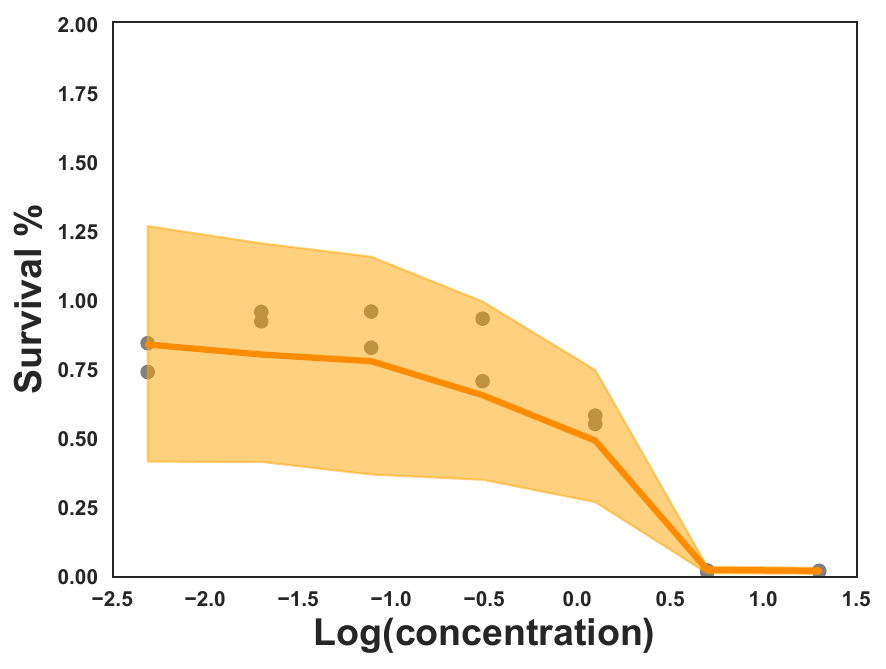}\caption{Organoid 1, Drug 2}\end{subfigure}
\begin{subfigure}{0.3\linewidth}\includegraphics[width=0.97\textwidth]{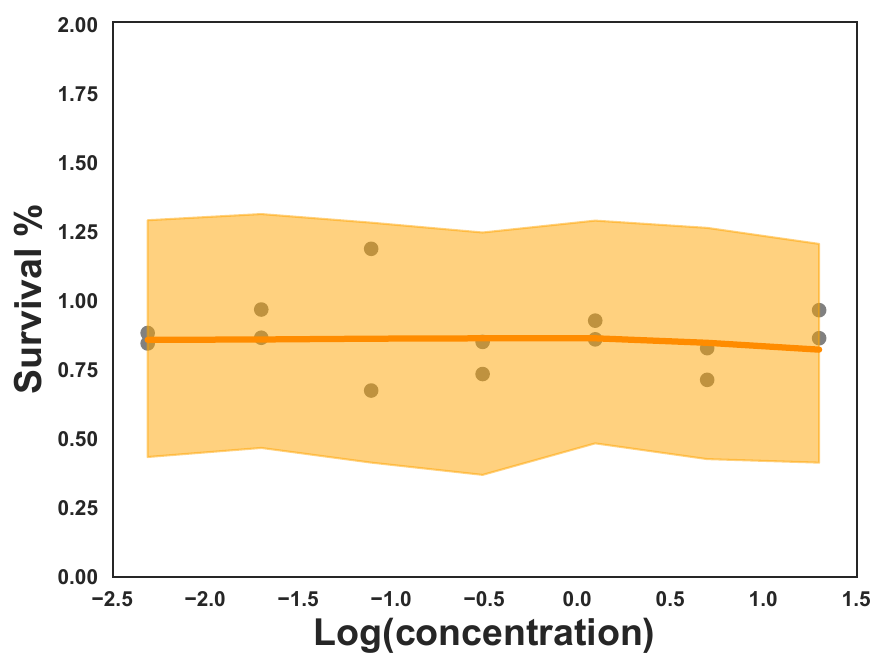}\caption{Organoid 1, Drug 3}\end{subfigure}
\begin{subfigure}{0.3\linewidth}\includegraphics[width=0.97\textwidth]{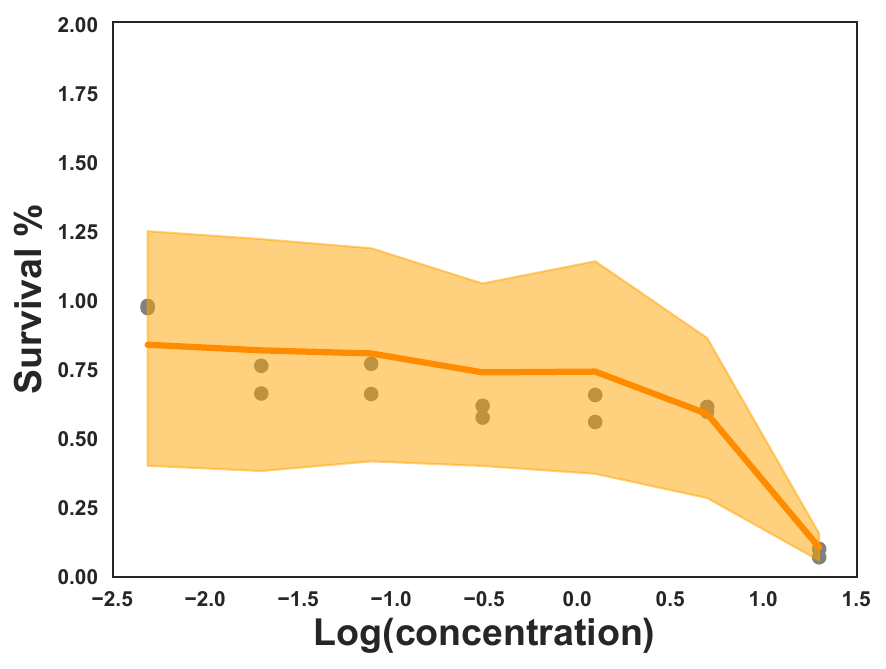}\caption{Organoid 2, Drug 1}\end{subfigure}
\begin{subfigure}{0.3\linewidth}\includegraphics[width=0.97\textwidth]{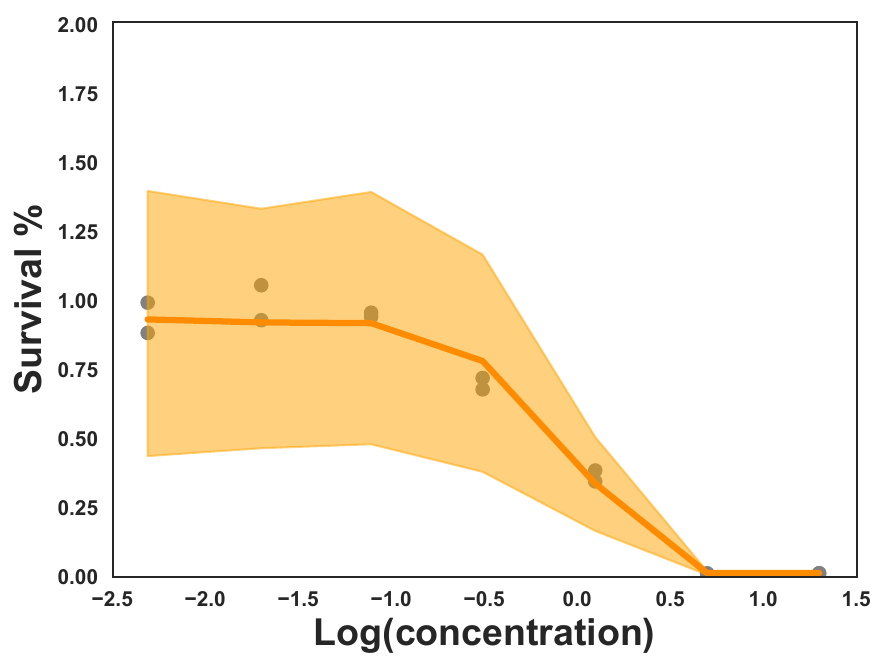}\caption{Organoid 2, Drug 2}\end{subfigure}
\begin{subfigure}{0.3\linewidth}\includegraphics[width=0.97\textwidth]{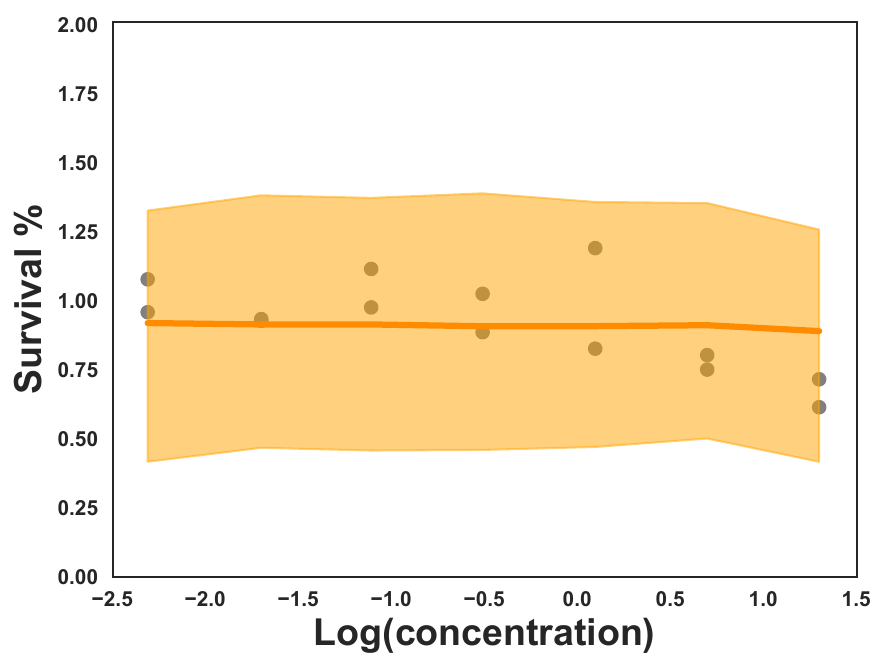}\caption{Organoid 2, Drug 3}\end{subfigure}
\begin{subfigure}{0.3\linewidth}\includegraphics[width=0.97\textwidth]{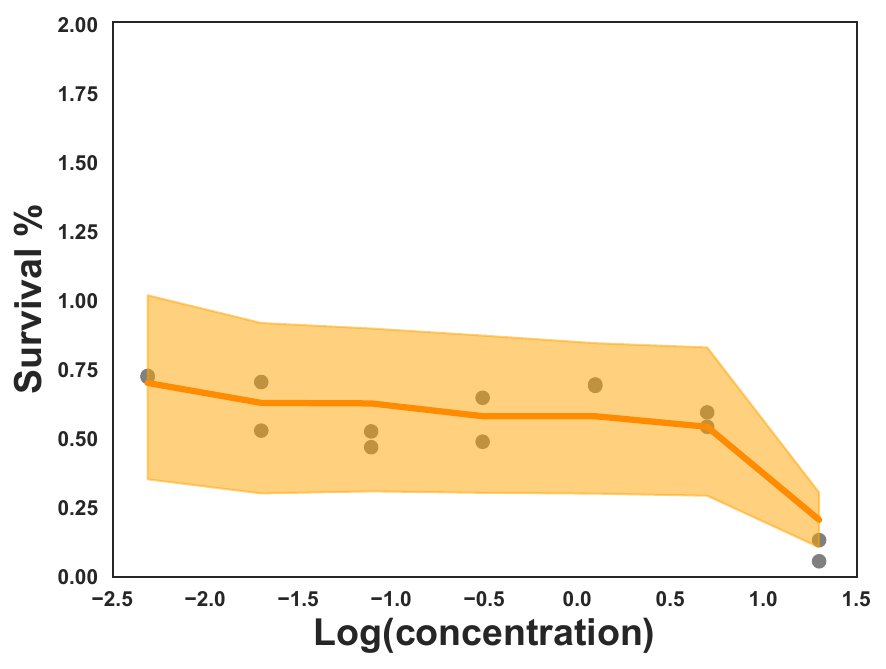}\caption{Organoid 3, Drug 1}\end{subfigure}
\begin{subfigure}{0.3\linewidth}\includegraphics[width=0.97\textwidth]{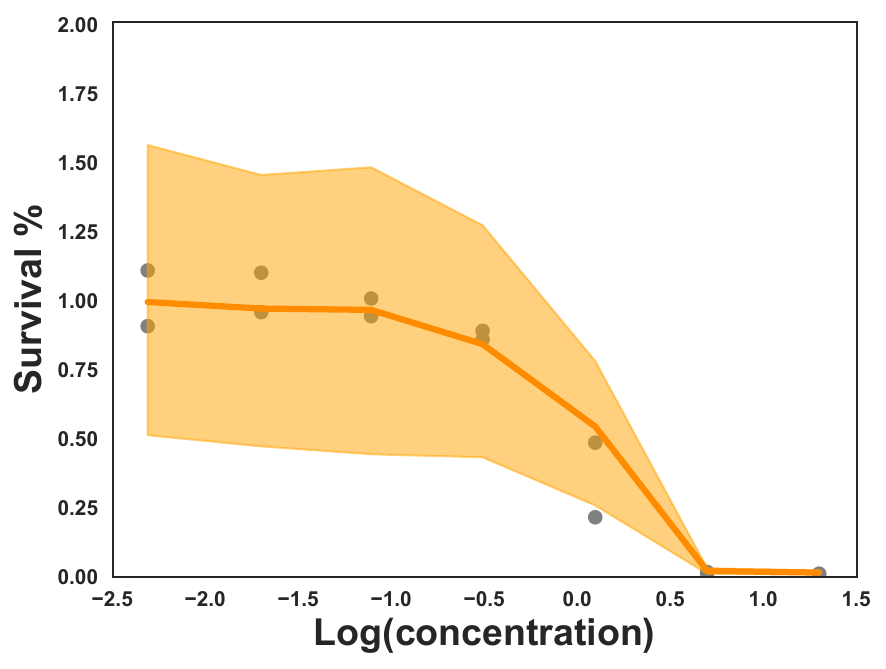}\caption{Organoid 3, Drug 2}\end{subfigure}
\begin{subfigure}{0.3\linewidth}\includegraphics[width=0.97\textwidth]{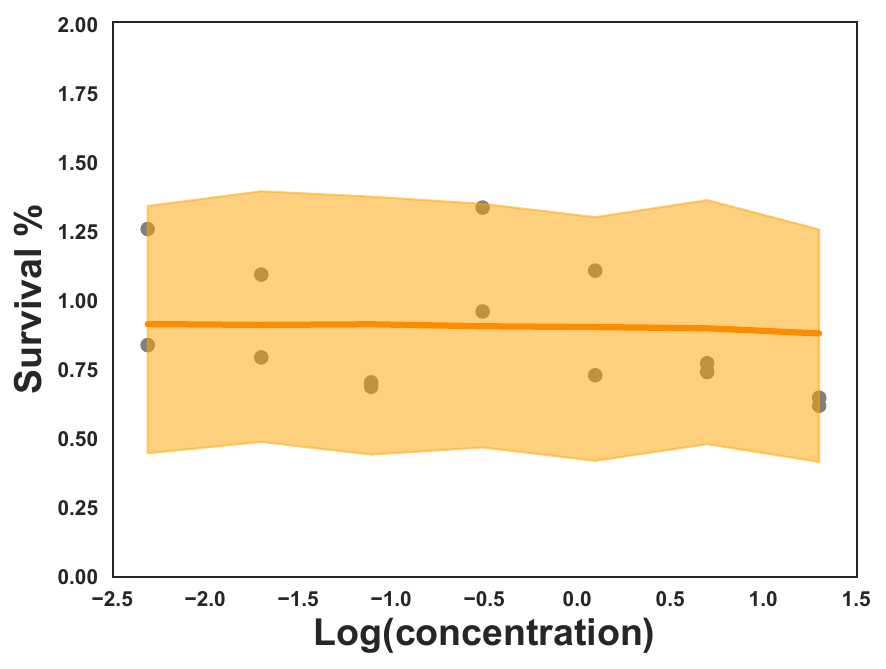}\caption{Organoid 3, Drug 3}\end{subfigure}
\caption{\label{fig:example} Sample of data from an organoid cancer drug experiment. Gray dots are observed outcomes; the orange line is the mean predicted response; bands represent $50\%$ posterior predictive credible intervals for observations. All nine experiments were held out from the model at training time.}
\end{figure}

\Cref{fig:example} shows data from the landscape study. 
Each panel
illustrates the interaction of one type of drug with one organoid sample. 
The gray points are the results of a set of
experiments, each set with $2$ replicates measured at $7$ different
doses.  The goal is to use the observations (gray points) to infer
the true dose-response curves. The predictions---the orange lines and uncertainty bands---come from
the Bayesian dose-response model we propose in this paper; each of the $9$ panels in \cref{fig:example} were
held out from the model at fitting time.





Notice there is structure in the outcomes: each drug has
similar effects on each organoid.  Thus, we treat modeling of
dose-response as a factorization problem. The structure in
the data arises because organoids share latent molecular attributes, such
as genomic mutations, and drugs share latent pharmaceutical attributes, such
as chemical structures. In each experiment, organoid and drug
attributes interact, creating the shared patterns of dose-response.

While traditional factorization considers a matrix of scalars, the
entries of this matrix are latent dose-response curves subsampled at
different doses. To model such curves, we model drug attributes
as
evolving with the dose level. While the effects usually vary
smoothly between dose levels, there are occasional sharp jumps, such as between the final two dose levels of drug $1$.
Capturing latent structure in
dose-response curves requires handling this type of non-stationarity.




The observation model for these experiments
is also non-standard. The outcome measured is a positive, real-valued measurement
of cell survival relative to a noisy baseline. The model we propose uses a
non-conjugate mixture-of-Gamma-shapes likelihood with the latent dose-response entering
through the
scale parameter. This is reflected in \cref{fig:example}, where the uncertainty intervals
shrink as the predicted survival rate drops.

As a final wrinkle, the drugs in these cancer studies are all cytotoxic, meaning that
they will only kill cells, not facilitate growth. 
This biological prior knowledge implies the expected survival rate
can only decrease as the dose increases. Cytotoxicity
adds a shape constraint requiring the dose-response curves are all monotonic.
Further, since these drugs will not facilitate growth, effects in each curve are also
upper bounded.


The dose-response model in this paper addresses these requirements.
First, we
propose Bayesian tensor filtering (BTF), a probabilistic method for smoothed tensor factorization.
BTF uses structured shrinkage
priors that encourage smoothness between successive dose levels, while
simultaneously enabling sharp jumps when the
data calls for it. Second, we develop generalized analytic slice sampling (GASS),
a new MCMC inference algorithm that, when combined with BTF,
enables the proposed dose-response model to support arbitrary
likelihoods and linear inequality constraints on dose-response curves. The
ability of GASS to handle linear constraints enables inference
of monotone dose-response curves with upper and lower bounds.

The remainder of this paper is structured as follows.  
We first review related work on Bayesian dose-response modeling in \Cref{sec:background}.
\Cref{sec:experiment_details} then provides an overview of the two cancer drug studies
and a motivation for the mixture-of-Gamma-shapes likelihood. 
\Cref{sec:model} presents Bayesian tensor filtering, a flexible model for smoothed tensor
factorization. In \Cref{sec:gass}, we detail generalized analytic slice sampling, a procedure
for sampling from posteriors with constrained multivariate normal
priors and non-conjugate likelihoods.
\Cref{sec:results} presents quantitative performance benchmarks for
BTF, GASS, and the proposed dose-response model.
Finally, in \Cref{sec:analysis} the dose-response model
is extended to handle side information in the form of molecular features about organoids.
An analysis of the landscape study dataset with $115$ features reveals
potential new biomarkers of drug sensitivity in a subset of organoids.



\section{Relevant literature}
\label{sec:background}
We survey a collection of the most relevant work to the proposed dose-response model. In each category of work, we focus on methods that enable uncertainty quantification, primarily through Bayesian inference. 



\paragraph{Bayesian factor modeling}
Many models have been developed for Bayesian factor analysis with smooth structure. \citet{zhang:paisley:2018:deep-bnp-tracking} apply a group lasso penalty to the rows and columns of a matrix then derive a variational expectation maximization (EM) algorithm \citep{bishop:2006:pattern-recognition} for inference. \citet{hahn:etal:2018:horseshoe-factor-model} use horseshoe priors \citep{carvalho:etal:2010:horseshoe} for sparse Bayesian factor analysis in causal inference scenarios with many instrumental variables. \citet{kowal:etal:2019:dynamic-shrinkage-processes} develop a time series factor model using a Bayesian trend filtering prior \citep{faulkner:minin:2018:bayesian-trend-filtering} on top of a linear dynamical system with P{\'o}lya--Gamma augmentation \citep{polson:scott:windle:2013:polya-gamma} for binomial observations. \citet{schein2:etal:016:pgds} develop Poisson-Gamma dynamical systems (PGDS), a dynamic matrix factorization model specifically for Poisson-distributed observations; we compare BTF with a tensor extension of PGDS in \Cref{sec:results}. Unlike the above models, BTF is likelihood-agnostic through GASS inference and enables modeling of independently-evolving columns rather than a common time dimension.

\paragraph{Independent dose-response curve estimation.}
A number of authors have investigated Bayesian methods for modeling monotone dose-response curves. These are typically done through a mixture of monotone functions. \citet{perron:mengersen:2001:bayesian-np-triangular-dose} use a mixture of triangular distributions. \citet{neelon:dunson:2004:bayesian-isotonic-regression} use an autoregressive mixture prior of truncated normals in a piecewise linear spline model. \citet{bornkamp:ickstadt:2009:bayesian-np-monotone-dose} propose a Bayesian nonparametric (BNP) model with a potentially-infinite mixture of two-sided power distributions. \citet{shively:etal:2009:bayesian-np-monotone-dose} also propose a BNP model which improves upon the model of \citet{neelon:dunson:2004:bayesian-isotonic-regression} with the key idea being to model the mean of the monotone curve as the integral of a positive function. \citet{ghebretinsae:etal:2013:bayesian-gamma-dose} present a Bayesian hierarchical model for nonnegative, real-valued comet assays with a Gamma outcome model on the shape. These methods all focus on the case of individual curve estimation. \citet{lin:dunson:2014:monotone-gp} propose a Gaussian process model with a posterior projection approach for shape-constrained curves. The datasets we consider here differ in that there are multiple samples and multiple drugs, with the goal to share statistical strength between samples and drugs to both denoise the existing curves and predict drug effects on samples without that specific (sample, drug) pair yet tested.

\paragraph{Joint dose-response curve estimation.}
A smaller body of work considers joint modeling of a set of dose-response curves. Both \citet{vis:etal:2016:logistic-dose-response} and \citet{abbas:etal:2019:nonlinear-mixed-effects-cancer} use nonlinear mixed effect models for cancer drug response in a regression setting. \citet{patel:etal:2012:bayesian-spline-dose} take a Bayesian B-spline approach to dose-response surface modeling. \citet{fridley:etal:2009:bayesian-loglinear-dose} propose a hierarchical Bayesian log-linear model for dose-response in cytotoxicity studies. \citet{fox:dunson:2015:bnp-covreg} consider the similar setting of covariance regression for influenza infection rates, imposing a sparse factorization on the covariance matrix that evolves over time in a Bayesian nonparametric setting. \citet{wilson:etal:2014:hierarchical-bayesian-dose} use monotone piecewise-linear splines in a hierarchical model of chemical toxicity assays, imposing a hierarchical Bayesian model that shrinks across similar molecules.

These models all have the common property that they shrink together samples via hierarchical priors, sharing statistical strength among rows to improve dose-response curve estimation. However, the hierarchical priors do not model any relational structure to shrink across samples and assays and do not provide any way to infer missing curves. Modeling the relational structure in multi-sample, multi-drug studies is crucial for cancer drug studies as often only a subset of samples have been tested for any given drug. Predictions about the missing curves can inform which experiments, among the many possible (sample, drug) combinations, show promise and should be carried out next. We discuss the deeper connections between BTF and both \citet{fox:dunson:2015:bnp-covreg} and \citet{wilson:etal:2014:hierarchical-bayesian-dose} after presenting the details of BTF; we also compare against \citet{fox:dunson:2015:bnp-covreg} with a Gaussian likelihood version of BTF in the benchmarks.

\paragraph{Predictive dose-response modeling.}
In many experiments, descriptive features are gathered representing useful side information about the samples or assays. A natural approach in these scenarios is to build predictive models that map from features to dose-response curves, enabling out-of-sample prediction for untested (sample, drug) pairs. \citet{low-kam:etal:2015:bart-dose-response} proposed a Bayesian regression tree model with spline leaf nodes, enabling prediction of entire dose-response curves from chemical descriptors in nanoparticle experiments. \citet{wheeler:2019:bayesian-additive-dose-response} modeled dose-response with molecular descriptors via additive Gaussian process tensor products over a real-valued feature space. In the case of binary or discrete data, \citet{wheeler:2019:bayesian-additive-dose-response} first take a principal components decomposition to project features to a continuous space, making feature interpretation difficult. Both methods also assume a Gaussian noise model with no shape constraints; dealing with non-conjugate likelihoods and shape constraints would require a novel inference scheme similar to the proposed GASS algorithm. \citet{tansey:etal:2018:deep-dose-response} use deep neural networks to predict monotone dose-response curves from molecular features in an approximate Bayesian model, but require a large dataset of experiments and features to train the neural network. More generally, predictive methods generally assume features to be available and complete. In the pilot dataset, no features are present; in the landscape dataset, many samples are missing feature information. In \cref{sec:analysis} we extend the dose-response model to include potentially-missing features via a multi-view factorization approach.

\section{Study design and dataset details}
\label{sec:experiment_details}
We detail the specific protocol used for the internal pilot study at Columbia University Medical Center. The landscape study we analyze uses a different number of plates, wells per plate, concentrations, drugs per plate, and replications. These distinctions are changes to the dimensions of the resulting tensor of observations, but the fundamental statistical inference challenges remain the same.

\begin{figure}[t]
\centering
\begin{subfigure}{0.47\linewidth}\includegraphics[width=\linewidth]{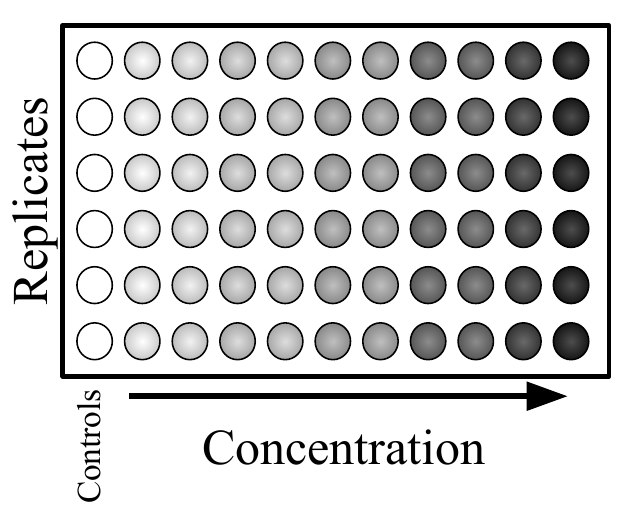}\caption{\label{fig:plate} Experimental design}\end{subfigure}
\begin{subfigure}{0.47\linewidth}\includegraphics[width=\linewidth]{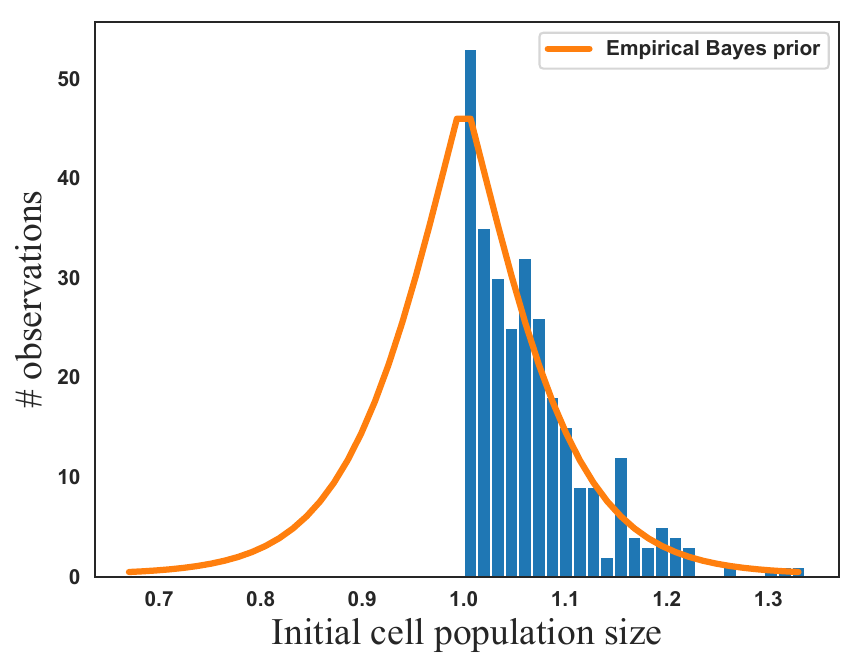}\caption{\label{fig:empirical-bayes} Empirical Bayes estimate}\end{subfigure}
\caption{\label{fig:dose-setup} Left: The layout of each microwell plate experiment used to generate a single dose-response experiment. Cells are pipetted one column at a time, leading to correlated errors. Right: Estimate of the prior distribution of mean cell counts in each column, relative to the control column mean. The prior is estimated empirically assuming the lowest concentration had no effect if it had a higher mean.}
\end{figure} 

Each dose-response experiment is conducted on a microwell plate where a single drug is tested against a single biological sample. Each experiment measures a proxy for cell abundance $72$ hours after applying the drug. Cell abundance is reported relative to a baseline control population where no drug was applied. For the control and each concentration level, $6$ replicates are tested. \Cref{fig:plate} shows the design of each $60$-well plate experiment in the pilot study. The proxy measurement is a fluorescence assay; a fluorescent protein is added after $72$ hours that binds to a molecule kept at near-constant levels in living cells. The degree of brightness of each microwell measures the relative abundance of cells alive but does not correspond to an exact cell count. All observations in the final dataset are normalized by dividing the brightness measurements in each microwell by the mean brightness for the control wells.

The details of how these organoid experiments were carried out in the wet lab are important, as they induce a particular form of correlated errors in the observations.
The first step in each experiment is to pipette an initial population of cells into each of the $60$ microwells on the plate. This is a time consuming process for the biologist, often taking hours to pipette a single plate. To speed up the plating process, biologists use a multi-headed pipette that enables them to simultaneously fill an entire column of each plate. This reduces the burden on the biologist, but comes at a cost: correlated errors.

When a biologist fills a microwell, they first draw a pool of cells into the pipette. Given the small volumes involved in laboratory experiments, the actual number of cells drawn can vary substantially on a relative basis. Using a multi-headed pipette transforms this variation into a hierarchical model: first a pool of cells is drawn into the pipette, then it is split among all the heads. The majority of the variation comes in the initial sampling, with small noise added in the splitting process. This has the unintended side effect of creating correlated errors between all microwells in a single column. The drug concentrations are replicated along the same column, leading to a fundamentally unanswerable question: was the drug particularly effective at this dose or was the initial sample of cells particularly small by chance?

In other cancer drug experiments, such as high-throughput cancer cell line experiments \citep{shoemaker:2006:nci60,garnett:etal:2012:gdsc,ghandi:etal:2019:ccle}, wells are filled using designs that avoid such correlated errors. In particular, these experiments typically use an orthogonal design, pipetting the drug replicates along rows and cells along columns. High-throughput screens exhibit their own correlated error pathologies, often referred to as batch effects. A number of techniques have been developed to preprocess high-throughput screening data and remove or limit batch effects \citep{johnson:etal:2007:batch-effects,leek:etal:2010:batch-effects,lachmann:etal:2016:hts-spatial-bias,mazoure:etal:2017:hts-spatial-bias,tansey:etal:2018:deep-dose-response}. These techniques account for the microenvironment similarities (e.g. temperature, humidity) that create spatial and temporal correlation between errors in wells nearby on the same plate, or run in the lab on the same day, in high-throughput screens.

Unfortunately, batch effect correction techniques are not applicable here. The correlation between wells is due to multi-headed pipetting, not similarities in the plate microenvironment between temporally- or spatially-related wells. Unlike smoothly-varying spatial batch effects, the experimental bias in the organoid datasets creates correlations between wells in the same column, but does not vary smoothly between columns. Drug concentrations are also varied across columns but constant across rows. This makes it impossible to disentangle the drug effect from the pipetting error without any assumptions.

\section{Generative dose-response model for cancer organoid studies}
To model the latent dose-response curves in the pilot and landscape studies we make two high-level design decisions. First, the unidentifiable experimental design necessitates making some assumption about the dose-response curves. We take an empirical Bayes approach, leveraging the fact that most drugs are ineffective at the smallest dose level. After specifying the likelihood, a hierarchical structure is imposed to share statistical strength between similar organoids and similar drugs. We propose a Bayesian hierarchical model that imposes data-adaptive smoothness between successive doses and shares statistical strength via latent, low-dimensional embeddings.

\subsection{Smallest-dose assumption and empirical Bayes likelihood estimation}
\label{subsec:dose-response:likelihood}
The correlated errors in the columns render the exact effects unidentifiable. Each column has two latent variables affecting the final population size of cells: a dose-level effect from the drug and an initial population size from the pipetting. Since both of these variables affect all replicates in a column, disentangling them precisely is impossible.

We take an empirical Bayes approach to disentangling the variation in drug effects from the technical error in pipetting. In most experiments, the lowest concentration tested is too small to have any effect on cell survival. We therefore make the assumption that any experiment where the mean of the control replicates is lower than the mean of the replicates treated at the lowest concentration has effectively two sets of control columns. This enables estimation of the variation between means and an empirical Bayes prior for the pipetting error.

Specifically, we form a histogram of all lowest-concentration means greater than the control mean on the same plate. We then fit a Poisson GLM with $3$ degrees of freedom to the histogram to estimate the prior probability that the mean of the initial population of cells was higher than the control mean. We assume the true distribution is symmetric and obtain an empirical Bayes prior on the means. \Cref{fig:empirical-bayes} shows an the histogram and empirical Bayes prior estimate for the pilot study dataset. The within-column variance is identifiable and estimated using the control replicates.

We integrate out the uncertainty in the initial population mean. For each organoid sample $i = 1, \ldots, N$ treated with drug $j = 1, \ldots, M$ at dose level $t = 1, \ldots, T$, with replicates $r = 1, \ldots, R$, this yields a gamma mixture model likelihood,
\begin{equation}
\label{eqn:empirical_bayes_likelihood}
P(y_{ijtr} \mid \mu_{ijt}) = \prod_{r=1}^{R} \left( \sum_{k=1}^{K} \hat{m}_k Ga(y_{ijtr}; \hat{a}_k, \hat{b}_k \mu_{ijt}) \right) \mathds{1}[0 \leq \mu_{ijt} \leq 1] \, ,
\end{equation}
where $(\hat{m}, \hat{a}, \hat{b})_k$ are derived from the empirical Bayes procedure. The latent drug effect $\mu_{ijt} \in [0,1]$ is the probability of a cell in organoid $i$ surviving treatment with drug $j$ at dose level $t$. The drug effect enters in the Gamma scale as the initial cell population size sampled from $ Ga(y_{ijtr}; \hat{a}_k, \hat{b}_k)$ is being multiplied by $\mu_{ijt}$. The drug effect is constrained to be a proportion, as the drugs are known not to help any cells grow (i.e., the proportion can be at most $1$) and a drug cannot kill more than all of the cells.

\subsection{Bayesian tensor filtering for organoid dose-response modeling}
\label{sec:model}
To capture shared structure between dose-response curves of similar organoid samples and similar drugs, we place smoothed factor model priors on the dose-response curve in a hierarchical Bayesian model we term Bayesian tensor filtering (BTF),
\begin{equation}
\label{eqn:generative_model}
\begin{aligned}
y_{ijtr}  &\sim& P(y_{ijtr} \mid \mu_{ijt}) \mathds{1}[\mu_{ijt} \leq \mu_{ij(t-1)}] \\
\mu_{ijt} &=& w_i^\top v_{jt} \\
w_i &\sim& \mathcal{N}_D(\mathbf{0}, \sigma^2 I_D) \\
(\Delta^{(k)} V_j)_\ell &\sim& \mathcal{N}_D(\mathbf{0}, \rho^2 \tau^2_{j\ell} I_D) \\
\tau_{j\ell} &\sim& \text{C}^+(0, \phi_{j\ell}) \\
\phi_{j\ell} &\sim& \text{C}^+(0, 1) \\
\sigma^{-2} &\sim& \text{Gamma}(0.1, 0.1) \\
\rho &\sim& P(\rho) \, .
\end{aligned}
\end{equation}
In the above model, $w_i, v_{jt} \in \R^D$ are the latent factors and loadings for each organoid $i$ and drug-dose $(j, t)$, respectively. The $T \times D$ drug loadings matrix $V_j$ contains all $(v_{j1}, \ldots, v_{jT})$ loading vectors for drug $j$. The choice of the number of latent factors, $D$, is a hyperparameter. We will occasionally refer to factors and loadings as embeddings or attributes, and $D$ as the embedding dimension.

The generative model in \cref{eqn:generative_model} incorporates a number of design decisions motivated by prior knowledge about biology and the nature of the experiments conducted. We explain the rest of the model in the context of these design decisions and the properties they induce in the resulting dose-response curves.

\paragraph{Monotonicity in the dose-response curve}
All drugs in both organoid studies are cytotoxic, only inducing a higher rate of cell death as the dose increases. This monotonic dose-response relationship is encoded by the hard constraint at the top of \cref{eqn:generative_model}. Each dose-response effect $\mu_{ijt}$ is required to be at least as toxic as the previous $\mu_{ij(t-1)}$ effect; we assume $\mu_{ij0} = 1$.

\paragraph{Latent attributes for biological samples}
Organoid samples share molecular attributes. In cancer, different tumor samples contain similar patterns of genomic mutations, copy number alterations, and gene expression \citep{weinstein:etal:2013:tcga}. In mixed tissue experiments, cells that have differentiated into the same type will often respond similarly \citep[e.g.,][]{huang:etal:2011:light-therapy-cell-types}. These attributes are captured in BTF with a latent vector, $w_i \in \R^D$ for the $i^{\text{th}}$ sample, as in standard matrix factorization. For identifiability \citep{bernardo:etal:2003:identifiable-bayesian-factor-models}, we assume a lower-triangular structure on the factors matrix $W = (w_1, \ldots, w_N)$, though interpretable factors is not our primary goal here.

\paragraph{Independent dose-specific latent attributes for drugs}
For each drug in the dataset, BTF models each dose level $t = 1, \ldots, T$ with its own embedding. For a single drug $j$, there are $T$ embeddings forming a drug embedding matrix $V_j \in \R^{T \times D}$. In BTF, columns (drug effects) are evolving independently, though potentially with similar latent attributes. This column independence distinguishes BTF from time-series tensor factorization models \citep{xiong:etal:2010:temporal-cf,spiegel:etal:2011:temporal-link-prediction,gauvin:etal:2014:temporal-community-structure,takeuchi:etal:2017:temporal-predictions} where all columns are progressing through time together. Independent column evolution captures the notion that two drugs treated at the same concentration may have totally different effects due to the molecular size of the drug, its targeting receptor, and its chemical structure. Different drugs may also be treated at entirely different concentration levels, as they have different molecular properties that require higher or lower concentration ranges to map out the dose-response relationship. For ease of notation, we assume all drugs are treated at $T$ concentrations, however conceptually drugs could be treated at different numbers $T_j$ of concentrations.

\paragraph{Non-stationary group smoothness priors on drug attributes}
We assume drug effects typically vary smoothly with dose, with occasional sharp jumps. To encode this, we place hierarchical smoothness priors on the differences between dose-specific drug embeddings. Specifically, we place a Bayesian group trend filtering prior on drug embeddings.

Trend filtering is an adaptive smoothing technique originally developed in the penalized regression case \citep{kim:etal:2009:trend-filtering,tibshirani:2014:trend-filtering}. The penalized regression formulation places $\ell_1$ penalties on the $k^{\mathrm{th}}$-order differences of neighboring points on a 1d grid. For instance, in the $k=1$ case this is the total variation norm penalty,
\begin{equation}
\label{eqn:trend_filtering}
\underset{\beta \in \R^n}{\mathrm{minimize}} \vnorm{Y - \beta}_2^2 + \lambda \vnorm{\beta_{1\mathord{:}(n-1)} - \beta_{2\mathord{:}n}}_1
\end{equation}
The solution to the convex optimization problem in \cref{eqn:trend_filtering} leads to piecewise constant plateaus in $\beta$ due to the lasso penalty driving first differences to zero; higher order differences lead to piecewise-polynomial solutions \citep{tibshirani:2014:trend-filtering}.

\citet{faulkner:minin:2018:bayesian-trend-filtering} extended trend filtering to the Bayesian context and considered a number of different priors on the $k^{\mathrm{th}}$-order differences. They consider three different priors on the differences: normal priors, equivalent to $\ell_2$ Laplacian smoothing in the regression case; Laplace priors, the direct Bayesian analog of $\ell_1$ lasso priors; and horseshoe priors \citep{carvalho:etal:2010:horseshoe}, a heavier tailed distribution that does not suffer from the non-diminishing bias of the lasso \citep{vanderpas:etal:2014:horseshoe-asymptotics}. The horseshoe priors are shown to perform best across a range of problems. We adapt the horseshoe prior to the group trend filtering case in BTF.

We call $\Delta^{(k)} \in \R^{L \times T}$ the composite trend filtering matrix; it contains all linear operators needed to encode the $(1, \ldots, k)^{\mathrm{th}}$-order differences. The ordinary trend filtering matrix encodes only the $k^{\text{th}}$-order differences, implicitly assuming all lower-order differences are not smooth. For example, the $k=2$ case yields a prior on the first and second order differences,
\begin{equation}
\label{eqn:tf_matrix}
\Delta^{(2)} = \left[\begin{matrix}
1 & 0 & 0 & 0 &  \ldots & 0 & 0 & 0 \\
1 & -1 & 0 & 0 &  \ldots & 0 & 0 & 0 \\
0 & 1 & -1 & 0 &  \ldots & 0 & 0 & 0 \\
&&&& \ldots &&& \\
0 & 0 & 0 & 0 &  \ldots & 0 & 1 & -1 \\
1 & -2 & 1 & 0 &  \ldots & 0 & 0 & 0 \\
0 & 1 & -2 & 1 &  \ldots & 0 & 0 & 0 \\
&&&& \ldots &&& \\
0 & 0 & 0 & 0 & \ldots & 1 & -2 & 1 \\
\end{matrix} \right] \, .
\end{equation}
The first line of \cref{eqn:tf_matrix} places an independent prior on the embedding vector for the first dose level in each drug, $v_{j1}$, making the matrix $(\Delta^\top \mathcal{T} \Delta)$ non-singular, where $\mathcal{T} = \mathrm{diag}(1/(\rho^2\tau^2_j))$. This ensures the resulting prior on $V_j$ is proper; see the supplementary material for details.

We impose a group trend filtering prior on the drug effects by placing a group horseshoe prior on the $\ell^{\mathrm{th}}$ row of the $(\Delta^{(k)} V_j)$ differences matrix. To do this, in \cref{eqn:generative_model} we adapt the Bayesian formulation of the group lasso \citep{kyung:etal:2010:bayesian-group-lasso} to the global-local shrinkage view of the horseshoe prior \citep{polson:scott:2010:shrink-globally}. Each row $(\Delta^{(k)} V_j)_{\ell}$ in the differences matrix has both a local $\tau_{j\ell}^2$ variance and a global $\rho^2$ variance term. Small values of $\rho^2$ and $\tau^2_{j\ell}$ will shrink the $\ell^{\text{th}}$ difference vector to nearly zero, resulting in the curve being smoother; larger values enable the curve to jump in response to the data.

Following \citet{bhadra:etal:2017:horseshoe-plus}, we place a half-Cauchy prior on $\phi_{j\ell}$, the scale term in the local horseshoe shrinkage prior; this is referred to as the horseshoe+ prior. A full Bayesian specification could choose a reasonable prior for $\rho$, such as a standard Cauchy or $\text{Uniform}(0,1)$. If an estimate of the number of non-zero entries is available, \citet{vanderpas:etal:2014:horseshoe-asymptotics} make an asymptotic argument for setting $\hat{\rho}$ to the expected number of non-zeros. We find BTF is robust to the choice of global shrinkage parameter; we default to a half-Cauchy prior on $\rho$ in our implementation and also support performing a grid search over a range of discrete $\rho$ values via deviance information criteria \citep{celeux:etal:2006:dic}. The value $k$ in $\Delta^{(k)}$ is left as a hyperparameter; we suggest $k=2$ as a reasonable default choice for most datasets. 


%

\subsection{Deeper connections to related work}
The BTF model is closely related to two works: the Bayesian nonparametric covariance regression model of \citet{fox:dunson:2015:bnp-covreg} and the zero-inflated piecewise log-logistic dose-response model of \citet{wilson:etal:2014:hierarchical-bayesian-dose}. Before proceeding to inference in BTF, we first provide a detailed discussion of the deeper connections to these methods.

\paragraph{Bayesian covariance regression.}
\citet{fox:dunson:2015:bnp-covreg} introduced Bayesian nonparametric covariance regression (BNP-CovReg). The BNP-CovReg model poses a Gaussian noise model for a set of $p$ curves observed at $n$ points,
\begin{equation}
\label{eqn:covreg-flu-basic-model}
\mathbf{y}_i = \boldsymbol\mu(\mathbf{x}_i) + \boldsymbol\epsilon_i \, , \quad \boldsymbol\epsilon_i \sim \mathcal{N}_{p}(0, \Sigma), \quad i = 1, \ldots, n\, ,
\end{equation}
where $\mathbf{y}_i = \log \mathbf{r}_i$, the vector of observations in the $p$ curves at point $x_i$, $\Sigma = \mathrm{diag}(\sigma_1, \ldots, \sigma_p)$. A factor model is imposed on the latent mean,
\begin{equation}
\label{eqn:covreg-flu-factor-model}
\mathbf{y}_i = \boldsymbol\Lambda(\mathbf{x}_i)\boldsymbol\eta_i + \boldsymbol\epsilon_i \, , \quad \boldsymbol\eta_i \sim \mathcal{N}_{k}(\boldsymbol\psi(\mathbf{x}_i), I_k)
\end{equation}
where $\boldsymbol\Lambda(\mathbf{x}_i)$ are the factor loadings at point $\mathbf{x}_i$ and $\boldsymbol\eta_i$ the latent factors associated with observation $\mathbf{y}_i$. Here, $k << p$ imposes a low-rank assumption on the factor model and independent Gaussian process priors are placed on each $\psi_h$, for $h = 1, \ldots, k$ with squared exponential kernel. Independent conjugate inverse-gamma priors are placed on each $\sigma_i$.

For computational feasibility, the factor loadings matrix $\boldsymbol\Lambda(\mathbf{x})$ is expressed as a weighted combination of a smaller set of $L$ basis functions,
\begin{equation}
\label{eqn:covreg-flu-basis-expansion}
\boldsymbol\Lambda(\mathbf{x}) = \boldsymbol\Theta \boldsymbol\xi(\mathbf{x}) \, ,
\end{equation}
where $\boldsymbol\Theta$ is a $p \times L$ matrix of coefficients and $\boldsymbol\xi(\mathbf{x})$ an $L \times k$ array of basis functions. A global-local shrinkage prior is placed on the elements of $\boldsymbol\Theta$ to effectively reduce the dimension of the basis to much smaller than $L$ or $k$.

A number of follow-up works have investigated similar models. \citet{kunihama:etal:2019:bayesian-np-longitudinal-covreg} extend BNP-CovReg to longitudinal data with covariate information. \citet{li:etal:2019:fixed-factors-covreg} use fixed factors, sacrificing the flexibility of the nonparametric approach of \citet{fox:dunson:2015:bnp-covreg} for increased scalability. \citet{heaukulani:vanderwilk:2019:variational-covreg} derive a variational inference approach for inverse Wishart processes that leads to a scalable approximate inference scheme for BNP-CovReg.

In the Gaussian likelihood case, BTF also uses a low-dimensional factor model for the response mean. However, rather than assuming independent sparsity and smoothness priors on a set of basis coefficients and latent factors, BTF imposes smoothness directly on the curves. This translates to a \textit{group} smoothness assumption on the latent factors, enforcing that the $k^{\mathrm{th}}$-order differences in successive latent means be shrunk to zero. As we show in \cref{subsec:results:flu-trends}, BTF outperforms the BNP-CovReg model on the same dataset that motivated the design of BNP-CovReg. It also enjoys substantially faster computational times (around 10x faster on a 2018 MacBook Pro) and trivial parallelization in the Gibbs sampler if further scalability is needed. Furthermore, BTF is extensible to a number of other likelihoods, including non-conjugate models with linear constraints like monotonicity, through the GASS inference algorithm.

\paragraph{Piecewise log-logistic, monotone dose-response modeling}
\citet{wilson:etal:2014:hierarchical-bayesian-dose} introduced the zero-inflated piecewise log-logistic (ZIPLL) model for estimating dose-response curves in chemical toxicity assays. Similar to the cancer drug study scenario, the ZIPLL model is explicitly designed for multi-sample, multi-assay studies where observations form a tensor with concentration as a third dimension. The ZIPLL model assumes a Gaussian noise distribution on observations $y_{ijt}$ of sample $i$, assay $j$, concentration $x_{ijt}$, respectively,
\begin{equation}
\label{eqn:zipll}
y_{ijt} = f_{ij}(x_{ijt}) + \epsilon_{ijt}\, , \quad \epsilon_{ijt} \sim \mathcal{N}(0, \sigma^2) \, .
\end{equation}
In the toxicity study considered, the minimum and maximum values corresponding to no-effect and total toxicity are unknown and modeled as latent variables,
\begin{equation}
f(x_{ijt}) = \begin{cases}
u_{ij} - (u_{ij} - l_{ij}) \times \mathrm{Logistic}(g(x_{ijt}; a_{ij}, \mathbf{w}_{ij})) & \mathrm{if } Z_{ij} = 1 \\
l_{ij} & \mathrm{if } Z_{ij} = 0
\end{cases} \, ,
\end{equation}
where $u_ij$ is the maximum response, $l_ij$ is the minimum response, $a_ij$ is the location parameter of the curve (the $50\%$ survival point), and $\mathbf{w}_{ij}$ are the shape parameters. The $Z_{ij}$ variable captures the zero-inflated property of the chemicals in the toxicity dataset, where sparsity in effects is expected. The ZIPLL model uses a monotonic log-linear spline basis to model $g(x_{ijt}; a_{ij}, \mathbf{w}_{ij})$ using a fixed symmetric grid of internal knots. Multivariate normal priors are placed on $\boldsymbol\theta_{ij} = \log (l_{ij}, u_{ij}, a_{ij})$ and a Gaussian autoregressive prior is placed on $\log \mathbf{w}_{ij}$ to encourage smoothness in the dose-response curve. Shrinkage in ZIPLL is performed across samples in the same assay via a hierarchical prior,
\begin{equation}
\label{eqn:zipll-shrinkage}
\boldsymbol\theta_{ij} \mid \boldsymbol\mu_j, \Sigma_j \sim \mathcal{N}_3(\boldsymbol\mu_j, \Sigma_j) \, , \quad \boldsymbol\mu_j \sim \mathcal{N}_3(\boldsymbol \mu, \Sigma) \, ,
\end{equation}
where hard-coded hyperparameters are used for conjugate priors on $\boldsymbol\mu$, $\Sigma_j$, and $\Sigma$, as well as the other hyperparameters in the hierarchical model.

The ZIPLL model is closely related to the BTF dose-response model, in both modeling constraints and target application domain. The use of a monotone basis with Gaussian priors on log-transformed latent parameters ensures the posterior curves satisfy the monotonicity constraints in cancer dose-response modeling, and the logistic transform maps to the $[0,1]$ interval as well. However, the Gaussian noise model is also misaligned with the cancer drug studies likelihood, and the stationary smoothness assumptions of the Gaussian autoregressive prior may not handle to sharp jumps found in the data. 

Moreover, ZIPLL only pools statistical strength across samples (rows) in the dose-response tensor but not assays (columns); this prevents imputing assays that are missing entirely. Extending the comparatively simple scalar basis coefficients $\mathbf{w}_{ijl}$ with autoregressive priors to a factor model with equivalent smoothness would be nontrivial, as posterior inference is non-conjugate and may require a novel approximate inference scheme. These are similar challenges to the inference in the BTF dose-response model, all of which led us to the development of the GASS algorithm in \cref{sec:gass}. The ability to impute out of sample experiments is critical for the cancer organoid drug studies, where only a subset of samples are tested for each drug.

\section{Posterior inference}
\label{sec:gass}

Posterior inference in BTF is performed through Gibbs sampling. In its most basic form, Gibbs sampling requires us to sample from the conditional distributions for each of the parameters. The updates for the latent attributes $W$ and $V$ depend on the form of likelihood, $P(y_{ijtr}; w_i^\top v_{jt})$. The derivations for Gaussian and binomial likelihoods, as well as the horseshoe parameter updates, are provided in the supplementary material. Here we focus on the crux of the posterior inference challenge in BTF: the constrained, non-conjugate likelihood in \cref{eqn:empirical_bayes_likelihood}.

To sample from the complete conditional for each of the latent attributes, we run MCMC-within-Gibbs -- running a separate Markov chain whenever a Gibbs step requires a sample from the conditional distributions of the latent attributes $w_i$ and $v_{jt}$. The challenge to this step is that the likelihood imposes hard constraints on the values of entries. The inner product must be a probability, requiring $w_i^\top v_{jt} \in [0,1]$. Each inner product must also be no greater than the previous inner product, requiring $w_i^\top (v_{jt} \leq v_{j(t-1)}) \leq 0$. Running a naive MCMC algorithm such as Metropolis Hastings within the Gibbs sampler is likely to have a high rejection rate and lead to poor mixing. Instead, we develop an MCMC algorithm that is capable of directly handling generic likelihoods and arbitrary linear constraints.


\subsection{Generalized analytic slice sampling}
Sampling from the conditional distributions of the latent attributes can be reduced to the problem of sampling from the posterior of a vector $x$ with a multivariate normal prior constrained by a set of linear inequalities,
\begin{equation}
\label{eqn:gass:setup}
x \sim P(y; x) \mbox{MVN}(x; \mu, \Sigma) \mathbb{I}[Dx \geq \gamma] \, .
\end{equation}
Note in \cref{eqn:gass:setup} we are describing a generic problem that is distinct from \cref{eqn:generative_model}. The variables $(y, x, \mu, \Sigma, D, \gamma)$ in \cref{eqn:gass:setup} are correspondingly generic variables that are distinct from those used in \cref{eqn:generative_model}.

A natural candidate for sampling from such a distribution is elliptical slice sampling \citep{murray:etal:2010:elliptical-slice-sampling}, an empirically successful exact MCMC method for sampling from distributions with arbitrary likelihoods and multivariate normal priors. Elliptical slice sampling builds on the idea of slice sampling \citep{neal:2003:slice-sampling}, an MCMC algorithm that, given a point $x$, samples a new point $x'$ by first drawing a value $u$ uniformly over the range $0$ up to the likehood of $x$, and then by drawing $x'$ uniformly from the set of points whose likelihood is at least $u$. Computing the set of points whose likelihood is above a certain threshold is infeasible in general, and thus some form of rejection sampling is required. Unfortunately, in high dimensions, these rejection rates will be very large.

However, when we have a multivariate normal prior, we may utilize the fact that the contours of equal probability on the multivariate normal distribution are elliptical regions. Elliptical slice sampling exploits this observation by sampling a point $v$ from the prior distribution and computing the ellipse $\{x \cos(\theta) + v \sin(\theta) : \theta \in [-\pi, \pi]\}$ containing both $x$ and $v$. Then, as in slice sampling, it samples a likelihood $u$ and then performs a form of rejection sampling to sample $x'$ uniformly from the set of points \emph{on the ellipse} whose likelihood is at least $u$. Note that when the likelihood is reasonably smooth, there will always be a reasonably large interval of points around $x$ on the interval above this likelihood, and we will not have too many rejections. However, when the likelihood has hard constraints, these regions can be very small and lead to high rejection rates.


To address this, we extend elliptical slice sampling to directly handle constrained multivariate normal priors. The approach, which we call generalized analytic slice sampling (GASS), is a natural extension of the analytic slice sampling procedure of \citet{fagan:etal:2016:analytic-slice-sampling} for truncated multivariate normals. The key difference is that the original analytic slice sampler only considered centered truncated multivariate normals with no likelihood component. Generalizing this procedure to handle the more general case in \cref{eqn:gass:setup} requires handling several edge cases.

\begin{algorithm}[t]
 \KwData{Valid current point $x$, mean $\mu$, covariance $\Sigma$, log-likelihood $\mathcal{L}$, constraints $(D,\gamma)$}
 \KwResult{MCMC sample from $P(x') \propto \exp(\mathcal{L}(x')) \mbox{MVN}(x'; \mu, \Sigma) \mathbb{I}[Dx' \geq \gamma]$}
 $t = \mathcal{L}(x) + \log \epsilon, \qquad \epsilon \sim U(0,1)$\;
 Sample proposal $v \sim \mbox{MVN}(v; \mathbf{0}, \Sigma)$\;
 Grid approximation $\mathcal{G} = \mbox{grid}(-\pi, \pi)$\;
 \ForEach{constraint $(d_i,\gamma_i) \in (D, \gamma)$}{
    $a = d_i^\top(x - \mu)$, $b = d_i^\top v$, $c = \gamma_i - d_i^\top\mu$\;
    \If{$a^2 + b^2 - c^2 \geq 0$ and $a \neq -c$}{
        Get $\theta_1, \theta_2$ as in \cref{eqn:gass:trig}\;
        \eIf{$a^2 > c^2$}{
            $\mathcal{G} = \mathcal{G} \bigcap [\theta_1, \theta_2]$\;
        }{
            $\mathcal{G} = \mathcal{G} \bigcap ([-\pi, \theta_1] \bigcup [\theta_2, \pi])$\;
        }
    }
 }
 Generate candidate samples $\mathcal{X} = \{ x' \colon x \cos(\theta_g) + v \sin(\theta_g) + \mu, \theta_g \in \mathcal{G} \}$\;
 Select uniformly from sufficiently likely candidates $\{ x' \colon \mathcal{L}(x') \geq t, x' \in \mathcal{X} \}$.
 \caption{\label{alg:gass} Generalized analytic slice sampling (GASS) for constrained MVN priors}
\end{algorithm}

\subsection{Algorithm} The full GASS procedure is presented in \Cref{alg:gass}. The idea of GASS is to note that the constraints can be pushed inside the proposal update. Namely, given an ellipse and a set of linear constraints, it is relatively straightforward to exactly compute their intersection, and thus to restrict proposals to that region.

To see why, consider the case where we have a single linear constraint requiring that the output point satisfies $d^\top x' \geq \gamma$. Then a valid angle $\theta$ must satisfy $a \cos \theta + b \, \sin \theta - c \geq 0$, where $a = d^\top (x-\mu)$, $b = d^\top (v-\mu)$, and $c = \gamma - d^\top\mu$. Basic trigonometry implies that the feasible range of $\theta$ is a subset of $[-\pi,\pi]$ whose boundary points are given by
\begin{equation}
\label{eqn:gass:trig}
\theta_1, \theta_2 \ = \ 2 \arctan \left(\frac{b \pm \sqrt{a^2 + b^2 - c^2}}{a + c} \right) \, .
\end{equation}
There are two edge cases where the entire ellipse is valid: (i) $(a^2 + b^2 - c^2) < 0$ and (ii) $a = -c$. In the first case, we trivially have $a^2 + b^2 < c^2$ and therefore $a \cos\theta + b \sin\theta > c$ for all $\theta$. In the second case, the only place the constraint boundary intersects the ellipse exactly at a single point of the ellipse and thus its selection has probability zero. For all other cases, the subset is determined based on the sign of $a^2 - c^2$. A positive sign indicates the quadratic in the inequality is concave and \cref{eqn:gass:trig} defines the boundaries of a contiguous region; a negative sign indicates convexity and thus the complement of the interval. 

\begin{figure}[t]
\centering
\begin{subfigure}{0.32\linewidth}\includegraphics[width=0.97\textwidth]{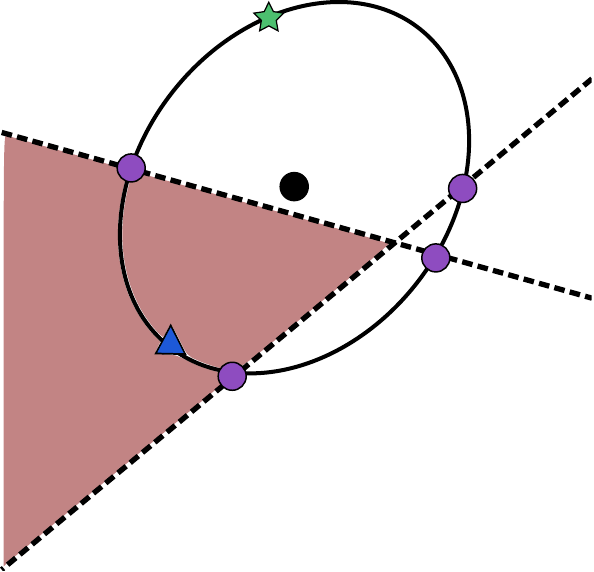}\caption{Sample ellipse}\end{subfigure}
\begin{subfigure}{0.32\linewidth}\includegraphics[width=0.97\textwidth]{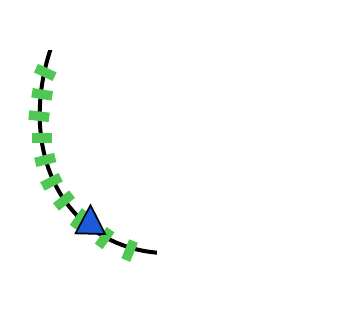}\caption{Compute valid interval}\end{subfigure}
\begin{subfigure}{0.32\linewidth}\includegraphics[width=0.97\textwidth]{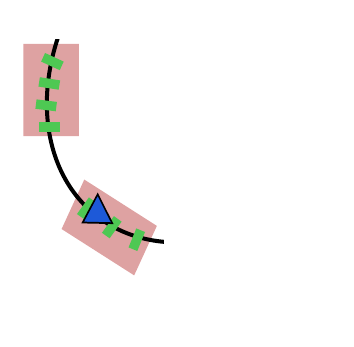}\caption{Compute likelihoods}\end{subfigure}
\caption{\label{fig:gass-illustration} An illustration of one step of the GASS chain starting from the blue triangle. (a) First, the green star is sampled form the multivariate normal centered at the black circle. The green star and the blue triangle determine an ellipse around the the black circle. Here the dashed lines denote the linear constraints, and the red region is the feasible region. (b) Then, the intersection of the feasible region with the ellipse is computed and gridded. (c) Finally, a likelihood $u$ is sampled, and those points that have likelihood at least $u$ are retained, denoted by the red region. The next state is randomly selected from the remaining grid points.}
\end{figure}

When there are many linear constraints, we can solve for the valid regions of each of the individual constraints separately and then take their intersection. Finally, after computing the region of valid $\theta$'s, we approximate it with a fine-grained 1D grid. We draw a likelihood value $u$ and filter out all the grid points with likelihood smaller than $u$. Finally, we draw a sample uniformly over the remaining grid points. An illustration of the algorithm is given in \cref{fig:gass-illustration}.
 
It is not difficult to show that GASS is a valid Markov chain which will converge to the distribution given in \cref{eqn:gass:setup}. For completeness, a proof of this convergence is provided in the supplementary material.
 
\subsection{Conditioning heuristic} Elliptical slice sampling schemes like GASS can suffer from poor mixing when the likelihood overwhelms the multivariate normal prior. In such settings, the sampled ellipses, which are generated with respect the prior and not the likelihood, may only have a small region centered around the current sample with likelihood comparable with the current likelihood, causing the chain to take only very small steps. Motivated by this observation, \citet{fagan:etal:2016:analytic-slice-sampling} suggest performing expectation propagation \citep{minka:2001:expectation-propagation} to better align the prior with likelihood. Unfortunately, in the context of BTF this is impractical as the prior parameters for $W$ are a function of $V$, and vice versa, which would require us to perform expectation propagation every iteration of the Gibbs sampler.


We instead approximate the entire tensor once at the start by a nonnegative, monotone tensor factorization. To do this, we find an approximate solution to the optimization problem,
\begin{equation}
\label{eqn:mono_nmf}
\begin{array}{lllll}
\hat{W}, \hat{V} &=& \underset{W,\mbox{ } V}{\mathrm{minimize}} && \sum_{ijtr} (y_{ijtr} - w_i^\top v_{jt})^2 \\
&& \mathrm{subject\mbox{ }to} && 0 \leq w_i^\top v_{jt} \leq 1\, , \\
&&&& w_i^\top (v_{jt} - v_{j(t-1)}) \leq 0
\end{array} \, ,
\end{equation}
The approximation solution to \cref{eqn:mono_nmf} is found via alternating constrained minimization for the rows and columns; we run the alternating minimization procedure until convergence.

After fitting the rows and columns, we calculate an over-estimate of the variance, analogous to an EP approximation, as a multiple of the empirical squared error in the estimate for each column and row,
\begin{equation}
\begin{aligned}
\hat{\mu}_{ijt} &=& \hat{w}_i^\top \hat{v}_{jt} \\
\hat{s} &=& \frac{\sum_{ijtr} (y_{ijtr} - \hat{\mu}_{ijt})^2}{N \times M \times T \times R} \\
\hat{\Sigma} &=& c \hat{s} I \, ,
\end{aligned}
\end{equation}
where $c \geq 1$ is a hyperparameter. This over-estimates the empirical variance, accounting for a wider range of possible samples to correct for the error in the NMF procedure. The main sensitivity of BTF inference is initialization and conditioning with an accurate mean; the covariance is less important. A reasonably accurate $\hat{\mu}$ yields the majority of the gains, whereas the method is insensitive to reasonable choices of $c$. As we discuss in the benchmarks, the NMF model produces a reasonably accurate $\hat{\mu}$ in terms of RMSE, and serves as a good starting point for the BTF dose-response model; for all BTF benchmarks, we set $c=3$.

The BTF dose-response model uses the conditioning heuristic in the $W$ and $V$ steps in the Gibbs sampler to calculate an adjusted prior. The log-likelihood used in the GASS procedure is then the original log-likelihood minus the log-conditioning likelihood, leaving the resulting distribution equivalent but better aligning the prior and likelihood.

\section{Benchmarks and performance comparisons}
\label{sec:results}
We study the performance of the proposed dose-response model and its components, BTF and GASS. We first benchmark GASS against different alternative methods for nonconjugate inference, where GASS mixes faster and has lower error. Then we study BTF on a dynamic matrix factorization problem with non-conjugate Poisson observations; BTF outperforms a recent Bayesian tensor decomposition approach designed for time-evolving count matrices. Finally, we apply the dose-response model to a real cancer drug study. We run $5$ independent trials, holding out a different subset of entire dose-response curves and report averages over all trials; the BTF-based dose-response model outperforms all baselines in terms of log probability on held out data.

\begin{table}
\centering
\begin{tabular}{lccccc}
\multicolumn{5}{c}{Google Flu Trends} \\ \hline
\multicolumn{1}{c|}{} & \multicolumn{2}{|c|}{In-sample} & \multicolumn{2}{|c}{Out-sample} \\ \hline
Model & MAE & RMSE & MAE & RMSE  \\ \hline
BNP-CovReg &  $0.31$ & $0.41$ & $0.31$ & $0.42$ \\
Gaussian BTF (d=2) & $0.15$ & $0.22$ & $0.15$ & $0.21$ \\
Gaussian BTF (d=5) & $0.10$ & $0.14$ & $\mathbf{0.12}$ & $\mathbf{0.16}$ \\
Gaussian BTF (d=10) & $\mathbf{0.07}$ & $\mathbf{0.10}$ & $0.13$ & $0.21$ \\ \\
\end{tabular}
\caption{\label{tab:flutrends_results} Posterior mean results on the Google Flu Trends dataset. The Gaussian BTF model outperforms the BNP-CovReg model of \citet{fox:dunson:2015:bnp-covreg} with as few as $2$ latent factors. MAE: mean absolute deviation from data; RMSE: root mean-squared error from data.}
\end{table}

\begin{figure}[t]
\centering
\begin{subfigure}{0.49\linewidth}\includegraphics[width=0.97\textwidth]{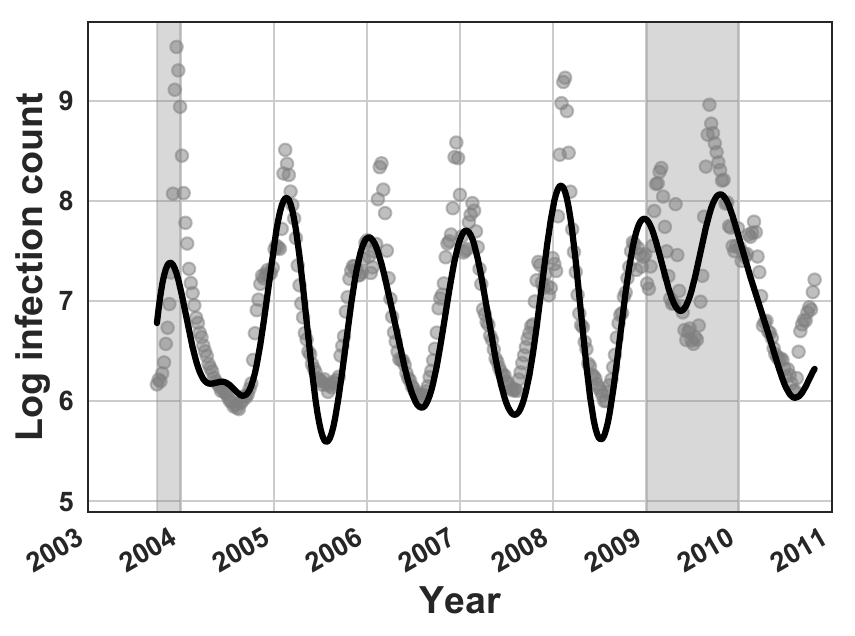}\caption{BNP-CovReg}\end{subfigure}
\begin{subfigure}{0.49\linewidth}\includegraphics[width=0.97\textwidth]{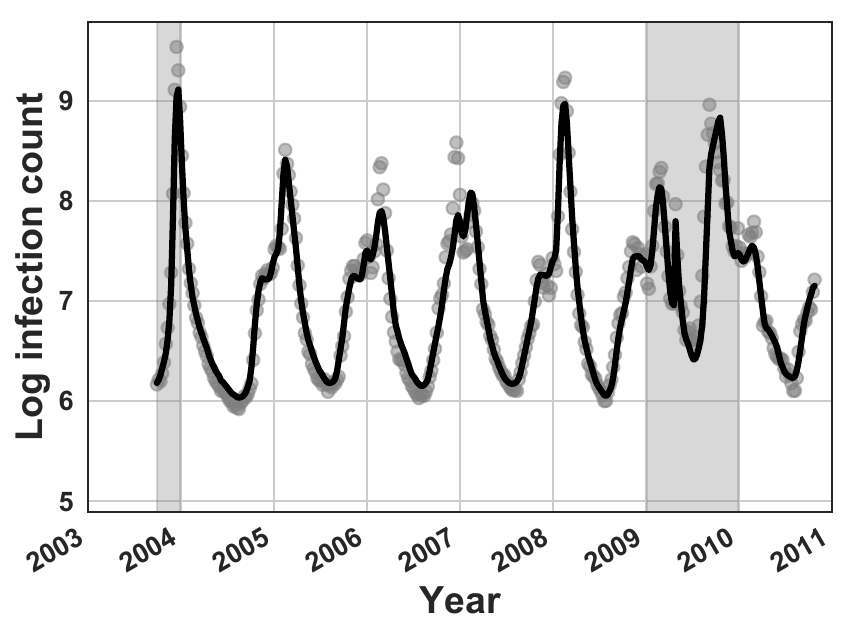}\caption{Gaussian BTF}\end{subfigure}
\caption{\label{fig:flutrends_benchmark} Example fits for BNP-CovReg and Gaussian BTF ($d=5$) on a single state (Alabama) in the Google Flu Trends dataset. The gray shaded regions represent held out periods. BNP-CovReg over-smooths, failing to capture large peaks and short movements. BTF fits the data tightly both on in-sample examples and imputed held out periods, suggesting it has captured latent structure between states.}
\end{figure}

\subsection{Gaussian BTF benchmarks}
\label{subsec:results:flu-trends}
We benchmark the conjugate Gaussian-likelihood BTF model on the Google Flu Trends dataset\footnote{\url{http://www.google.org/flutrends/}} modeled in \citet{fox:dunson:2015:bnp-covreg}. This dataset contains weekly influenza-like infection (ILI) counts from $183$ different regions in the United States from $1996$ to $2014$. The set of regions contains nested information, including both major cities and entire states. For benchmarking purposes, we focus only on the $50$ states to ensure that held out data is not leaked in through other nested regions; this makes the inference task strictly more difficult.

\citet{fox:dunson:2015:bnp-covreg} model the weekly log-count of infections with the Gaussian noise model in \cref{eqn:covreg-flu-basic-model}--\cref{eqn:covreg-flu-factor-model}, where $\mathbf{y}_i = \log \mathbf{r}_i$ is the vector of log Google-estimated ILI rates in the $50$ states at time $x_i$.

We compare against the performance of the Bayesian Nonparametric Covariance Regression (BNP-CovReg) model of \citet{fox:dunson:2015:bnp-covreg}, designed specifically for the Google Flu Trends data. For BNP-CovReg, we keep the same hyperparameter settings with truncation parameters $\bar{L}=10$ and $\bar{k}=20$; we use the reference implementation provided by the authors. For Gaussian BTF, we initialize the global shrinkage parameter $\rho^2$ to $0.1$ and sample it with a HS+ prior; we also place an weakly-informative inverse-$\mathrm{Gamma}(0.1, 0.1)$ prior on the likelihood variance. To measure performance, we hold out $10\%$ of all years, selected uniformly at random across all available state-years. Model performance is measured in both root mean squared error (RMSE) and mean absolute error (MAE) on held out data.

\Cref{tab:flutrends_results} shows the results for both the BNP-CovReg model and Gaussian BTF with $d=2$, $5$, and $10$ latent factors. The BTF method outperforms BNP-CovReg in each case. The model performs best out of sample with $d=5$ latent factors, suggesting it overfits with $d=10$ factors. The BTF model also has good coverage, with $95\%$ credible intervals having $95.83\%$ coverage on in-sample data and $92.82\%$ coverage for held out data with $d=5$ factors.

\Cref{fig:flutrends_benchmark} shows an example of a single state (Alabama) comparing BNP-CovReg and $d=5$ BTF. The BNP-CovReg model over-smooths, leading it to underestimate large peaks both in- and out-of-sample. The BTF model closely tracks the data, even in out-of-sample predicted weeks, suggesting it has learned latent structure between the states. 

\begin{figure}[t]
\centering
\begin{subfigure}{0.31\linewidth}\includegraphics[width=0.97\textwidth]{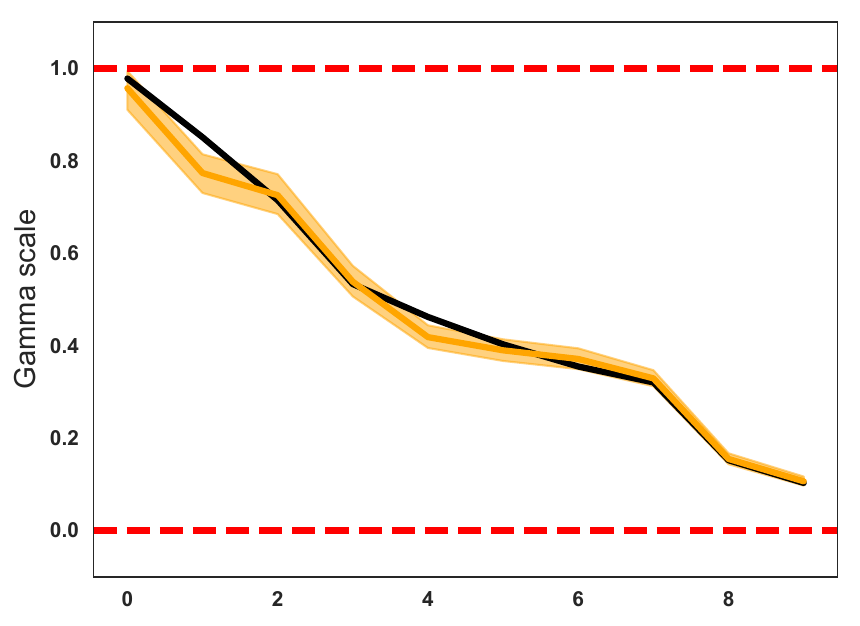}\caption{RS}\end{subfigure}
\begin{subfigure}{0.31\linewidth}\includegraphics[width=0.97\textwidth]{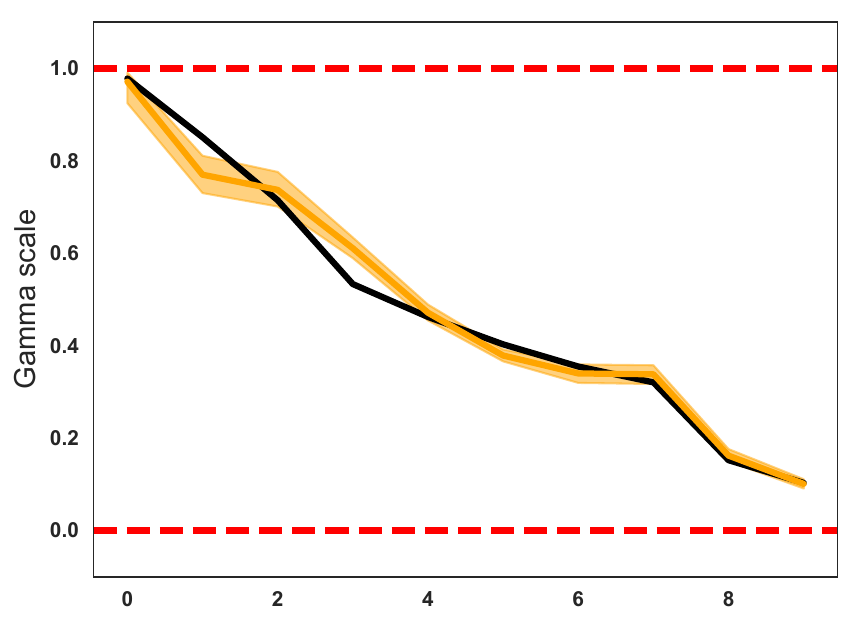}\caption{LRS}\end{subfigure}
\begin{subfigure}{0.31\linewidth}\includegraphics[width=0.97\textwidth]{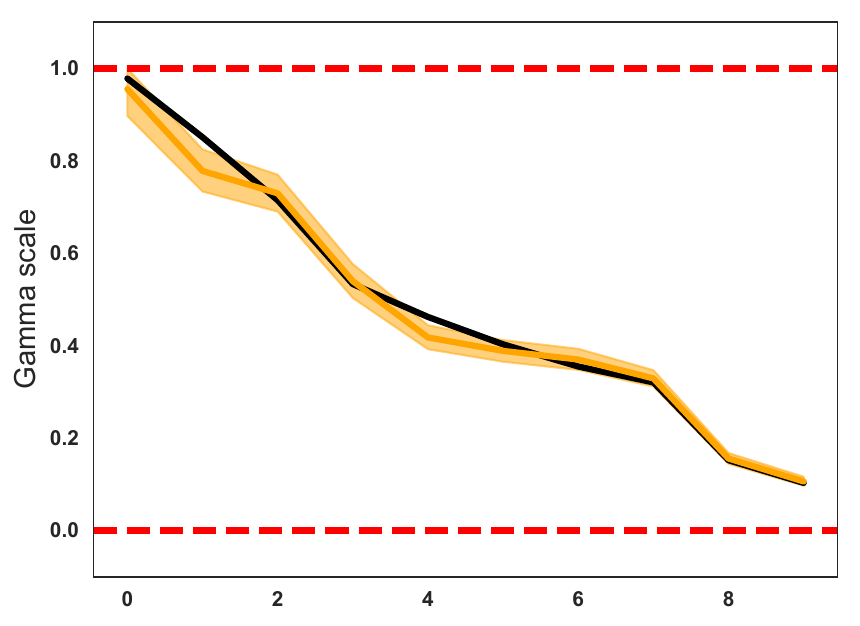}\caption{PP}\end{subfigure}
\begin{subfigure}{0.31\linewidth}\includegraphics[width=0.97\textwidth]{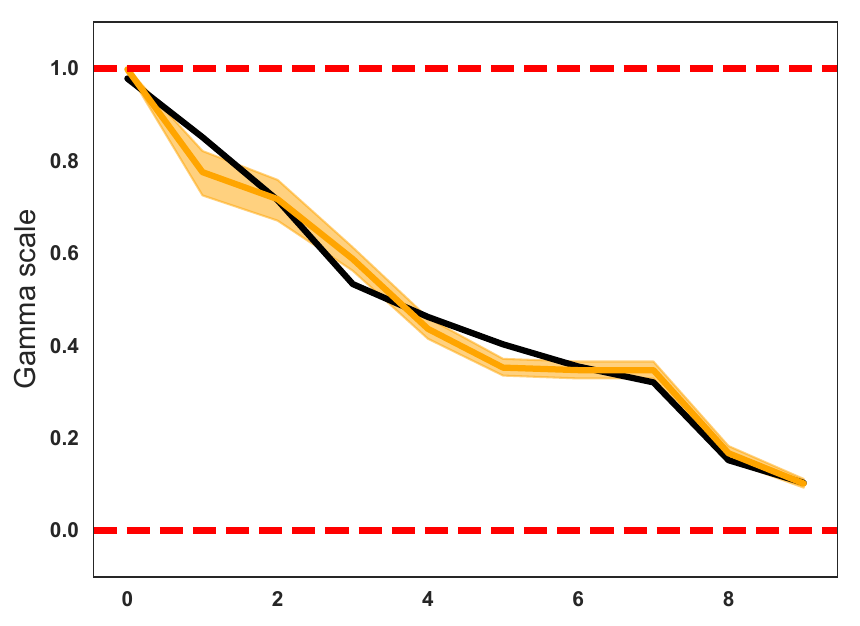}\caption{LPP}\end{subfigure}
\begin{subfigure}{0.31\linewidth}\includegraphics[width=0.97\textwidth]{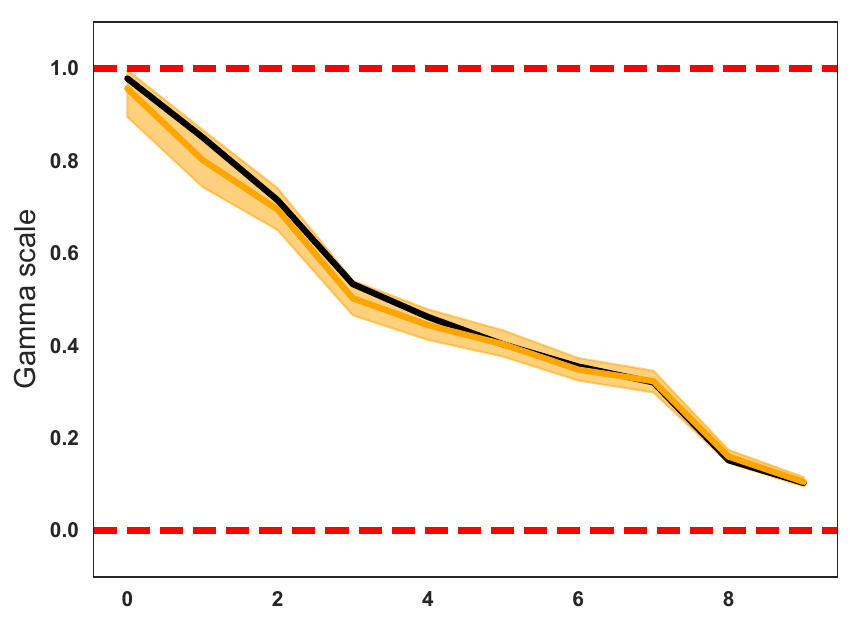}\caption{GASS}\end{subfigure}
\caption{\label{fig:gass_benchmark} Sample fits for different methods on the gamma scale estimation benchmark; black is the true scale, orange is the estimated scale, bands are $90\%$ credible intervals, dashed lines are constraint boundaries. GASS captures the shape of the curve and has good coverage after $10$K Gibbs iterations.}
\end{figure}

\begin{table}
\centering
\begin{tabular}{lccccc}
& \multicolumn{4}{c}{MSE ($\times 10^3$)}  \\ \hline
Sampler & $m=100$ & $m=500$ & $m=1000$ & $m=5000$ & $m=10000$ \\ \hline
RS & $1.54 \pm 0.09$ & $1.43 \pm 0.09$ & $1.43 \pm 0.09$ & $1.41 \pm 0.09$ & $1.41 \pm 0.09$ \\
LRS & $1.76 \pm 0.14$ & $1.75 \pm 0.14$ & $1.73 \pm 0.14$ & $1.66 \pm 0.13$ & $1.63 \pm 0.13$ \\
PP & $1.35 \pm 0.09$ & $1.31 \pm 0.08$ & $1.29 \pm 0.09$ & $1.29 \pm 0.09$ & $1.28 \pm 0.08$ \\
LPP & $1.66 \pm 0.12$ & $1.62 \pm 0.12$ & $1.56 \pm 0.12$ & $1.52 \pm 0.13$ & $1.50 \pm 0.12$ \\
GASS & $\mathbf{0.74 \pm 0.05}$ & $\mathbf{0.66 \pm 0.04}$ & $\mathbf{0.63 \pm 0.04}$ & $\mathbf{0.52 \pm 0.03}$ & $\mathbf{0.49 \pm 0.03}$ \\
\\
& \multicolumn{4}{c}{$90\%$ Credible Interval Coverage} \\ \hline
Sampler & $m=100$ & $m=500$ & $m=1000$ & $m=5000$ & $m=10000$ \\ \hline
RS & $0.37 \pm 0.02$ & $0.47 \pm 0.02$ & $0.49 \pm 0.02$ & $0.50 \pm 0.02$ & $0.52 \pm 0.02$ \\
LRS & $0.23 \pm 0.01$ & $0.34 \pm 0.02$ & $0.36 \pm 0.02$ & $0.46 \pm 0.02$ & $0.48 \pm 0.02$ \\
PP & $0.44 \pm 0.02$ & $0.54 \pm 0.02$ & $0.56 \pm 0.02$ & $0.57 \pm 0.02$ & $0.58 \pm 0.02$ \\
LPP & $0.30 \pm 0.01$ & $0.40 \pm 0.02$ & $0.44 \pm 0.02$ & $0.53 \pm 0.01$ & $0.53 \pm 0.02$ \\
GASS & $\mathbf{0.58 \pm 0.02}$ & $\mathbf{0.73 \pm 0.02}$ & $\mathbf{0.77 \pm 0.02}$ & $\mathbf{0.86 \pm 0.01}$ & $\mathbf{0.87 \pm 0.01}$
\end{tabular}
\caption{\label{tab:gass-benchmarks} Benchmark performance for GASS versus alternative non-conjugate elliptical sampling approaches. Results are averages over $100$ independent trials $\pm$ standard error.}
\end{table}

\subsection{GASS benchmarks}
We benchmark GASS on a simulation study with a constrained multivariate normal prior with a non-conjugate gamma scale likelihood,
\begin{equation}
\label{eqn:benchmarks:gass}
\begin{aligned}
y^{(r)}_i &\sim \mathrm{Gamma}(y_i; a, \theta_i) \\
\boldsymbol\theta &\sim \mathrm{MVN}(\boldsymbol\theta; \boldsymbol\mu, \Sigma) \mathds{1}[0 \leq \boldsymbol\theta \leq 1] \prod_{i=1}^{n-1}\mathds{1}[\theta_i \geq \theta_{i+1}] \\
\boldsymbol\mu &= [0.95, 0.8, 0.75, 0.5, 0.29, 0.2, 0.17, 0.15, 0.01, 0.0001] \\
\Sigma_{ij} &= \tau \exp(-\frac{1}{2b} (i - j)^2) \, .
\end{aligned}
\end{equation}
The covariance matrix in the unconstrained prior corresponds to a squared exponential kernel. We set the hyperparameters $a=100$, $\tau=0.1$, and $b=3$; all hyperparameters are assumed known. We use $R=3$ replicates for $\mathbf{y}$. We compare GASS against four different variants of elliptical slice sampling (ESS),
\begin{itemize}
    \item \textbf{Rejection sampling (RS).} Samples are drawn using the unconstrained ESS model, with the constraints pushed into the likelihood. Any violated constraint generates a zero probability and corresponds to a rejection sampler.
    \item \textbf{Logistic rejection sampling (LRS).} The ESS is used to model logits, which are then passed through the logistic transform to satisfy the $[0,1]$ constraint; rejection sampling again handles the monotonicity constraint.
    \item \textbf{Posterior projections (PP).} No constraints are imposed on the model during posterior inference. Instead, we use the posterior projection approach of \citet{lin:dunson:2014:monotone-gp} to post-hoc enforce the constraints.
    \item \textbf{Logistic posterior projections (LPP).} A hybrid of the previous two approaches combined: modeling the logits for $[0,1]$ constraints but projecting the posterior samples to the monotone surface.
\end{itemize}
All models use the true prior mean and covariance; for the logistic models, we empirically estimate the covariance of the logit-transformed $\theta$. We compare performance with $2m$ MCMC steps, where the first $m$ are a burn-in phase and the last $m$ are used for posterior approximation; we consider $m=[100,500,1000,5000,10000]$. Performance is measured in terms of mean squared error (MSE) and coverage rate of the $90\%$ credible intervals for every $\theta_i$ point. Results are averaged over $100$ independent trials with $(\boldsymbol\theta, \mathbf{y})$ resampled from \cref{eqn:benchmarks:gass} at the start of each trial.

\Cref{fig:gass_benchmark} shows examples of the fits for each method, with $90\%$ credible intervals. GASS is the only procedure that results in good coverage of the true mean and captures the shape of the overall curve after $20$K MCMC steps; the other methods tend to over-smooth the curve and underestimates the uncertainty.

\Cref{tab:gass-benchmarks} shows the aggregate results of the benchmarks. GASS outperforms all four comparison methods in terms of both error and coverage. After $m=100$ samples the MSE for GASS is lower and coverage is higher than any of the other strategies after $m=10000$ samples. Further, the model appears to have almost fully mixed after $5000$ samples, with the coverage rate close to $90\%$. 

\begin{table}
\centering
\begin{tabular}{lcccccccc}
\multicolumn{9}{c}{d=2}\\ \hline
\multicolumn{1}{c|}{} & \multicolumn{3}{|c|}{Observations} & \multicolumn{2}{|c|}{True Rate} & \multicolumn{3}{c}{Coverage} \\ \hline
Model & MAE & RMSE & NLL & MAE & RMSE & $50\%$ & $75\%$ & $95\%$ \\ \hline
NMF             & $1.70$              & $2.47$              & $364.29$            & $1.12$              & $1.72$              & N/A              &     N/A          &  N/A\\
PGDS(0.25)      & $1.76$              & $2.47$              & $360.76$            & $1.29$              & $2.00$              & $15.32$             & $25.35$             & $40.59$ \\
PGDS(0.5)       & $1.74$              & $2.43$              & $359.96$            & $1.31$              & $2.03$              & $14.43$             & $23.96$             & $38.94$ \\
PGDS(1)         & $1.74$              & $2.43$              & $360.29$            & $1.30$              & $2.02$              & $14.58$             & $23.93$             & $39.09$ \\
NBinom BTF      & $>10^2$            & $>10^3$              & $>10^5$             & $48.10$             & $>10^2$             & $30.17$             & $47.80$             & $69.67$ \\ \hline
Poisson BTF     & $\mathbf{1.66}$     & $\mathbf{2.34}$     & $\mathbf{354.52}$   & $\mathbf{0.91}$     & $\mathbf{1.43}$     & $\mathbf{34.16}$    & $\mathbf{53.40}$    & $\mathbf{74.72}$ \\
\\
\multicolumn{9}{c}{d=3}\\ \hline
\multicolumn{1}{c|}{} & \multicolumn{3}{|c|}{Observations} & \multicolumn{2}{|c|}{True Rate} & \multicolumn{3}{c}{Coverage} \\ \hline
Model & MAE & RMSE & NLL & MAE & RMSE & $50\%$ & $75\%$ & $95\%$ \\ \hline
NMF             & $2.26$              & $3.48$              & $552.10$            & $1.15$              & $1.88$              & N/A              &     N/A          &  N/A\\
PGDS(0.25)      & $1.72$              & $2.42$              & $365.26$            & $1.06$              & $1.68$              & $21.58$             & $35.83$             & $56.24$ \\
PGDS(0.5)       & $\mathbf{1.71}$     & $2.42$              & $364.39$            & $1.06$              & $1.68$              & $21.51$             & $35.58$             & $55.69$ \\
PGDS(1)         & $\mathbf{1.71}$     & $\mathbf{2.41}$     & $363.86$            & $1.05$              & $1.68$              & $21.27$             & $35.00$             & $54.81$ \\
NBinom BTF      & $>10^3$             & $>10^4$             & $>10^6$             & $>10^2$             & $>10^3$             & $37.14$             & $59.20$             & $82.60$ \\ \hline
Poisson BTF     & $1.78$              & $2.47$              & $\mathbf{358.95}$   & $\mathbf{0.79}$     & $\mathbf{1.29}$     & $\mathbf{40.45}$    & $\mathbf{62.11}$    & $\mathbf{82.87}$ \\
\\
\multicolumn{9}{c}{d=5}\\ \hline
\multicolumn{1}{c|}{} & \multicolumn{3}{|c|}{Observations} & \multicolumn{2}{|c|}{True Rate} & \multicolumn{3}{c}{Coverage} \\ \hline
Model & MAE & RMSE & NLL & MAE & RMSE & $50\%$ & $75\%$ & $95\%$ \\ \hline
NMF             & $2.65$              & $4.21$              & $596.37$            & $1.43$              & $2.31$              & N/A              &     N/A          &  N/A\\
PGDS(0.25)      & $1.74$              & $2.44$              & $368.10$            & $0.86$              & $1.37$              & $31.46$             & $50.20$             & $73.55$ \\
PGDS(0.5)       & $1.76$              & $2.43$              & $369.59$            & $0.87$              & $1.37$              & $31.35$             & $50.46$             & $73.83$ \\
PGDS(1)         & $1.80$              & $2.56$              & $371.89$            & $0.88$              & $1.40$              & $29.80$             & $49.05$             & $71.91$ \\
NBinom BTF      & $>10^2$             & $>10^2$             & $>10^4$             & $10.77$             & $94.94$             & $\mathbf{45.31}$    & $\mathbf{67.85}$    & $\mathbf{89.66}$ \\ \hline
Poisson BTF     & $\mathbf{1.71}$     & $\mathbf{2.26}$     & $\mathbf{355.75}$   & $\mathbf{0.81}$     & $\mathbf{1.25}$     & $41.64$             & $63.55$             & $83.98$ \\ 
\\
\multicolumn{9}{c}{d=10}\\ \hline
\multicolumn{1}{c|}{} & \multicolumn{3}{|c|}{Observations} & \multicolumn{2}{|c|}{True Rate} & \multicolumn{3}{c}{Coverage} \\ \hline
Model & MAE & RMSE & NLL & MAE & RMSE & $50\%$ & $75\%$ & $95\%$ \\ \hline
NMF             & $4.53 $             & $7.45 $             & $>10^3$             & $1.89$              & $3.12 $             & N/A              &     N/A          &  N/A\\
PGDS(0.25)      & $1.86 $             & $2.74 $             & $391.60 $           & $0.82$              & $1.33 $             & $38.25$             & $60.21$             & $83.84$ \\
PGDS(0.5)       & $1.89 $             & $2.80 $             & $402.63 $           & $\mathbf{0.79}$     & $1.29 $             & $39.20$             & $61.48$             & $84.40$ \\
PGDS(1)         & $1.79 $             & $2.52 $             & $376.62 $           & $\mathbf{0.79}$     & $\mathbf{1.26} $    & $39.21$             & $61.52$             & $84.61$ \\
NBinom BTF      & $31.87$             & $42.16$             & $>10^3$             & $3.13$              & $11.14$             & $43.33$             & $66.17$             & $\mathbf{87.55}$ \\ \hline
Poisson BTF     & $\mathbf{1.72} $    & $\mathbf{2.43}$     & $\mathbf{359.12} $  & $0.83$              & $1.32 $             & $\mathbf{44.05}$    & $\mathbf{67.03}$    & $86.59$ \\
\end{tabular}
\caption{\label{tab:poisson-benchmark} Mean results on the Poisson dynamical system benchmark; smaller is better for all metrics. NMF: nonnegative matrix factorization; PGDS($\tau$): Poisson-gamma dynamical system with hyperparameter $\tau$; NBinom-BTF: Bayesian tensor filtering with conditionally-conjugate negative binomial likelihood; Poisson-BTF: Bayesian tensor filtering with constrained, non-conjugate Poisson likelihood via GASS inference; NLL: negative log-likelihood; MAE: mean absolute deviation from truth; RMSE: root mean-squared error from truth. Bold indicates the best performance at each embedding dimension setting.}
\end{table}

\subsection{Non-stationary Poisson dynamical systems}
\label{subsec:results:poisson}
We benchmark BTF on a synthetic Poisson tensor dataset where the observations are Poisson distributed with a latent rate curve for each function. The rate at every point in the curve is the inner product of two gamma random vectors,
$$
\label{eqn:pgds:likelihood}
\begin{aligned}
h_{j\ell} \sim \mbox{Bern}(0.2) \, , &\mbox{ }& u_{j\ell d} \sim (1-h_{j\ell})\delta_{0} + h_{j\ell}\mbox{Ga}(1, 1) \, , &\mbox{ }& v_{jtd} = \sum_{\ell=1}^t u_{j\ell d}  \, ,\\
w_{id} \sim \mbox{Ga}(1, 1)  \, , &\mbox{ }& y_{ijt} \sim \mbox{Pois}(\langle w_i, v_{jt} \rangle) \, .
\end{aligned}
$$
The resulting true rates form a monotonic curve of constant plateaus with occasional jumps. As in the dose-response data, the columns evolve independently of each other, rather than through a common time parameter. We set the latent factor dimension to $3$.

We compare a Poisson likelihood version of BTF with GASS inference (Poisson BTF) to nonnegative matrix factorization (NMF), the Poisson-Gamma dynamical system (PGDS) model of \citet{schein2:etal:016:pgds}, and a negative binomial likelihood version of BTF (NBinom BTF) with {{P}}{\'o}lya--{{G}}amma augmentation \citep{polson:scott:windle:2013:polya-gamma}. For PGDS, we contacted the authors who suggested we try three different values of the hyperparameter $\tau=(0.25, 0.5, 1)$. For NBinom BTF, we use MCMC-within-Gibbs and sample the latent rate parameter with $30$ steps of random walk Metropolis-Hastings for every Gibbs step. For Poisson BTF, we initialize with NMF and did not use any conditioning heuristic.

We run all models for $5000$ burn-in iterations and collect $5000$ samples on an $11 \times 12 \times 20$ tensor with the upper left $3 \times 3 \times 20$ corner held out. We conduct $5$ independent trials, regenerating new data each time and evaluating the models on the held out data and true latent rate. We measure performance in three categories of metrics: (i) mean absolute error (MAE), root mean squared error (RMSE), and negative log-likelihood (NLL) on held out observations; (ii) MAE and RMSE on the true latent rate; and (iii) posterior credible interval coverage of the true rate at $50\%$, $75\%$, and $90\%$ targets. We evaluate all models at embedding dimensions $d=(2,3,5,10)$ to compare sensitivity to the common hyperparameter. 

\Cref{tab:poisson-benchmark} presents the results. The Poisson BTF model performs similarly across the range of dimension embeddings, whereas the other models are more sensitive. The NMF model generally performs better with a smaller embedding dimension while the PGDS model performs better with larger dimensions. In the case of NMF, this is due to overfitting without any smoothness prior built in. PGDS uses a canonical tensor decomposition, where the third tensor dimension is modeled with an embedding that is shared between rows and columns. This requires an inflation of the embedding dimension when columns are all evolving independently in order to sufficiently capture all of the latent structure in the data.

The negative binomial BTF model performs poorly on held out observations and true rate estimation. This is due to the instability of the model on held out data. The {{P}}{\'o}lya--{{G}}amma approach to negative binomial likelihoods uses the $NB(r, \sigma(\beta))$ parameterization, where $r$ is an unknown dispersion parameter, $\sigma$ is the logistic function, and $\beta$ are log-odds. The mean of the distribution is
$$
\frac{n \sigma(\beta)}{1 - \sigma(\beta)} \, .
$$
When the failure rate $\sigma(\beta)$ is near zero or one, small changes in $\beta$ lead to large changes in the mean. Thus, small errors in $\beta_{ijt} = w_i^\top v_{jt}$ on the held out data lead to large errors in the observational NLL and the MAE and RMSE metrics we consider.

For each embedding dimension, the Poisson BTF model performs competitively or better than the other models in each category. The negative log-likelihood on held out observations is always lowest for the Poisson BTF model. For other metrics, Poisson BTF is either the best performing model or within $10\%$ of the best performing model. By contrast, the other models are off by more substantial amounts in certain categories at certain dimension embedding settings. The credible intervals for Poisson BTF also consistently have better coverage than PGDS for the same embedding dimension. This includes the true embedding dimension $d=3$, where Poisson BTF shows the best performance in terms of MAE to the true rate, and competitive performance with other choices of $d$. By contrast, both NMF and PGDS perform substantially worse in the $d=3$ regime than in other choices. Finally, unlike the negative binomial model, the Poisson BTF model maintains a stable prediction on held out entries.

\begin{table}
\centering
\begin{tabular}{lccc}
\multicolumn{3}{c}{Cancer Drug Studies} \\ \hline
\multicolumn{1}{c|}{} & \multicolumn{1}{|c}{Pilot Study} & \multicolumn{1}{|c}{Landscape Study} \\ \hline
Model & NLL & NLL \\ \hline
NMF & $262.75 \pm 308.12$ & $25573.14$ \\
LMF & $589.17 \pm 582.29$ & $> 10^6$ \\
BTF & $\mathbf{-80.22 \pm 9.67}$ & $\mathbf{-3268.11}$ \\ \\
\end{tabular}
\caption{\label{tab:dose_results} Left: mean results $\pm$ standard error on held out data for the pilot cancer drug studies. Right: Results on a single test set of $1000$ curves for the landscape study. NMF: nonnegative matrix factorization; LFM: logistic factor model; BTF: Bayesian tensor filtering; NLL: negative log-likelihood.}
\end{table}

\subsection{Cancer drug study}
\label{subsec:results:cancer}
We evaluate the proposed empirical Bayes dose-response model, built on top of BTF, on two cancer drug studies. First, we use a small internal pilot study conducted at Columbia University Medical Center. The pilot study tested $35$ drugs against $28$ tumor organoids, each at $9$ different concentrations with $6$ replicates. Second, we run on a large-scale, ``landscape'' study \citep{lee:etal:2018:gbm-organoids-drug-response} that tested $67$ drugs against $284$ tumor organoids, each at $7$ different concentrations with $2$ replicates. For the pilot study, we run $5$ independent trials, holding out $30$ curves at random, subject to the constraint that no column or row is left without any observations in the training set. For the landscape study, we hold out a single test set of $1000$ curves ($\approx 5\%$ of the total entries). Since this is real data, MAE and RMSE from the truth are not available; we measure performance solely in terms of negative log-likelihood on the held out data.

The standard dose-response modeling approach in cancer datasets is a log-linear logistic model \citep{vis:etal:2016:logistic-dose-response}. For a baseline, we extend that model to a logistic factor model (LFM), using the same preprocessing strategy. We also compare to NMF as a second baseline. To ensure the monotonicity, we project the NMF results to be monotone curves using the PAV algorithm as in \citet{lin:dunson:2014:monotone-gp}.
We choose the factor size in both models by $5$-fold cross-validation on the training set.

For BTF, we perform a grid search over hyperparameters: $\rho^2 = \{0.001, 0.01, 0.1\}$, factor size $D=\{1,3,5,8\}$, and the order of the trend filtering matrix $k=\{0,1\}$; we select the best model using the deviance information criterion \citep{celeux:etal:2006:dic}. We evaluate the BTF model using the average of the posterior draws, rather than the full Bayes estimate; this enables us to fairly compare with the NMF and LFM point estimates of the latent mean. We run $10000$ Gibbs sampling steps in both studies, discarding the first $5000$ as burn-in.

\Cref{tab:dose_results} present the results. The BTF dose-response model outperforms both baselines in terms of negative log-likelihood of the held out data in both studies. Results in terms of RMSE and MAE (not shown) on the raw observations were similar for all three models in both studies (e.g. RMSE $0.14 \pm 0.01$, MAE $0.20 \pm 0.01$ in the pilot study). The BTF procedure is also more stable in the pilot study cross-validation, with a much lower reconstruction variance than either baseline. This suggests BTF not only forms a more accurate basis for a dose-response model, but is also more reliable.

Qualitative results on the held out predictions are in \Cref{fig:example} (orange). All $9$ plots are for real data from the landscape study, with the gray observations held out. The orange line shows the posterior mean of the predicted curves. The curves have all of the desired properties: monotonicity with dose, bounded between zero and one, mostly smooth, locally adaptive to sharp jumps in the data, and highly predictive of the outcomes of the experiments. The orange bands show the $50\%$ approximate posterior credible intervals using the empirical Bayes likelihood model. The credible intervals are conservative, estimating a larger variance than is actually observed in the outcomes. Even still, the NMF and LMF models far exceed these bands in certain points in the curve. This is due to the heteroskedastic nature of the likelihood and the misspecification of the NMF and LMF loss functions. Both competing models optimize for squared error, effectively making a heteroskedastic assumption on the model. In RMSE terms, all three models perform nearly identically, within $\pm 0.01$ of each other on both datasets. Judging the models by RMSE would be misleading, since the high degree of noise in the first three dose levels dominates the overall loss and obscures the real fit of the model.


\section{Landscape study analysis}
\label{sec:analysis}
In many cancer drug studies, molecular features such as gene expression, genomic mutations, or copy number alterations are gathered. These features represent useful side information that can help further denoise the dose-response data. Features also enable one to address the ``cold-start'' problem, enabling predictions for samples that have no dose-response data available. Features that are predictive of sensitivity or resistance in the dose-response experiments may represent ``biomarkers''-- diagnostic indicators of drug response. Biomarkers are candidate targets for future experimental investigation such as targeted drug development. The landscape study of \citet{lee:etal:2018:gbm-organoids-drug-response} considered $115$ molecular features, all binary, with a subset of the features gathered on a subset of organoid cell lines.

To incorporate potentially-missing features into the BTF dose-response model, we take a multi-view factorization approach. For each feature $x_{im}$, $m = 1, \ldots, M$, for organoid cell line $i$, we assume a latent factor model between the cell line embeddings and feature embeddings,
\begin{equation}
\label{eqn:feature-factors}
\begin{aligned}
x_{im} \mid w_i, u_m &\sim& \mathrm{Bern}(w_i^\top u_m) \mathds{1}[0 \leq w_i^\top u_{m} \leq 1] \\
u_m &\sim& \mathrm{MVN}(\mathbf{0}, \sigma_u^2 I) \, .
\end{aligned}
\end{equation}
Rather than a logistic link function with an unconstrained likelihood model, we use an identity link with a $[0,1]$ constraint. A logistic link would put the feature likelihood and dose-response likelihoods on different scales. By using a constrained identity link function in \cref{eqn:feature-factors}, a change in $w_i$ has the same effect on the probability of dose-specific survival as in the probability of a feature being positive.

For the landscape study, we set $\sigma_u = 1$ and use a latent embedding dimension of $10$. All other hyperparameters and settings are the same as in the benchmarks in \cref{subsec:results:cancer}.

\begin{figure}[t]
\centering
\includegraphics[width=\textwidth]{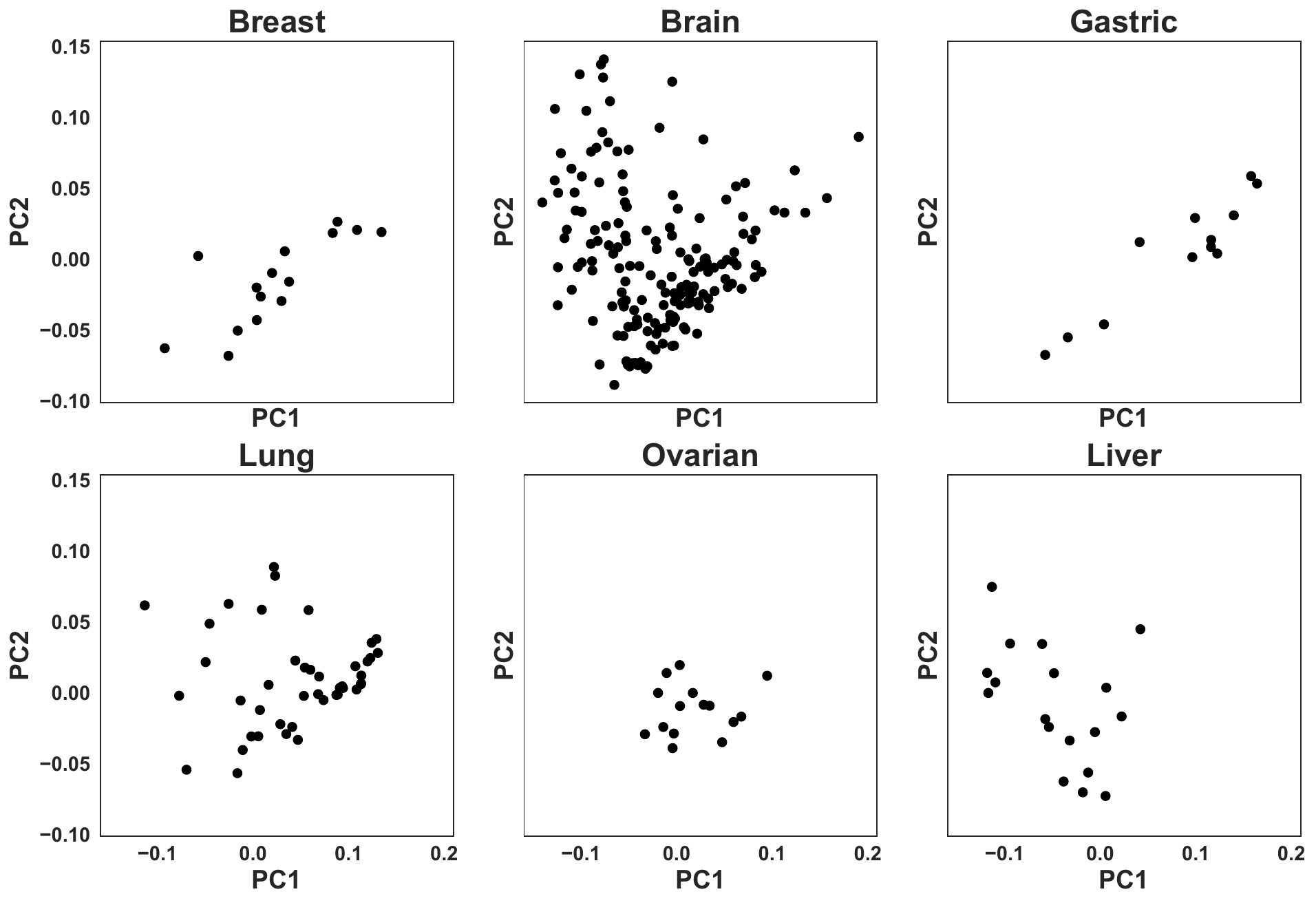}
\caption{\label{fig:embeddings} Two-dimensional PCA projection of the learned organoid embeddings, stratified by tumor site of origin. The embeddings reveal learned structure relating to site of origin, which was not an explicit input to the model.}
\end{figure}
\subsection{Site of origin structure captured by organoid embeddings}
\label{subsec:analysis:structure}
We first check whether the model has uncovered latent structure in the data. Tumor molecular profiles and drug response are strongly associated with the site of origin of the tumor. The BTF dose-response model was not supplied any direct site-of-origin information. Nonetheless, we expect that some clustering of organoids into site of origin should emerge.

Visualizing structure in the $10$-dimensional organoid embeddings is challenging, as with any data of more than $4$ dimensions. \Cref{fig:embeddings} shows a $2$-dimensional principal components analysis (PCA) of the posterior mean of the $10$-dimensional organoid embeddings. The $2$-d projection of the embeddings confirms that site of origin is associated with the first two principal components. Ovarian cancers are predominantly clustered in the bottom-center of the $2$-d space, liver cancers skew left, and breast and gastric cancers skew right. Brain cancers, cover the entire space but also represent the largest and most diverse set of original tumors in the landscape dataset.

\begin{figure}[t]
\centering
\begin{subfigure}{0.49\linewidth}\includegraphics[width=\textwidth]{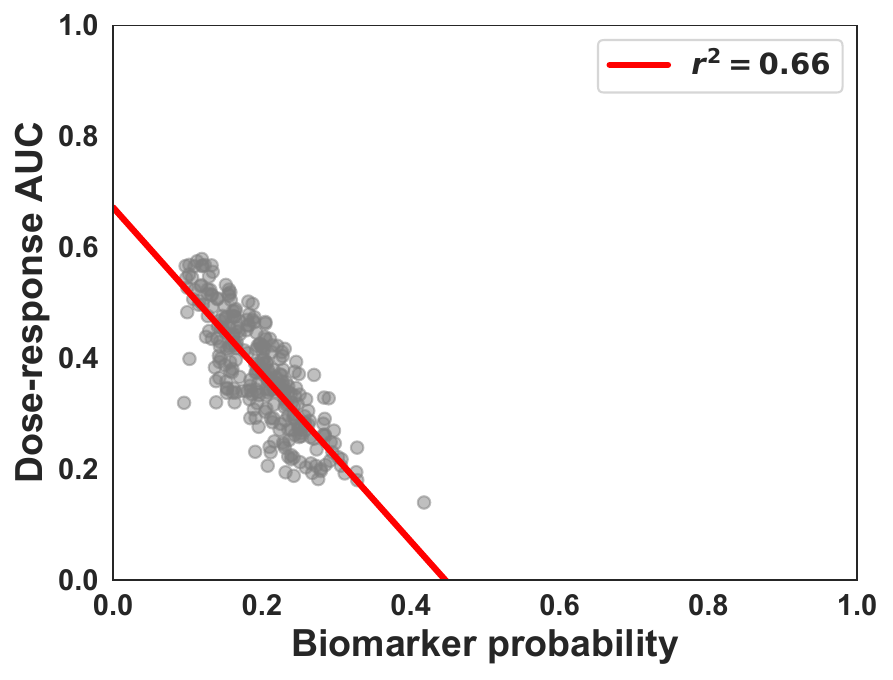}\caption{\label{fig:gene-drug:egfr}EGFRvIII + Trametinib}\end{subfigure}
\begin{subfigure}{0.49\linewidth}\includegraphics[width=\textwidth]{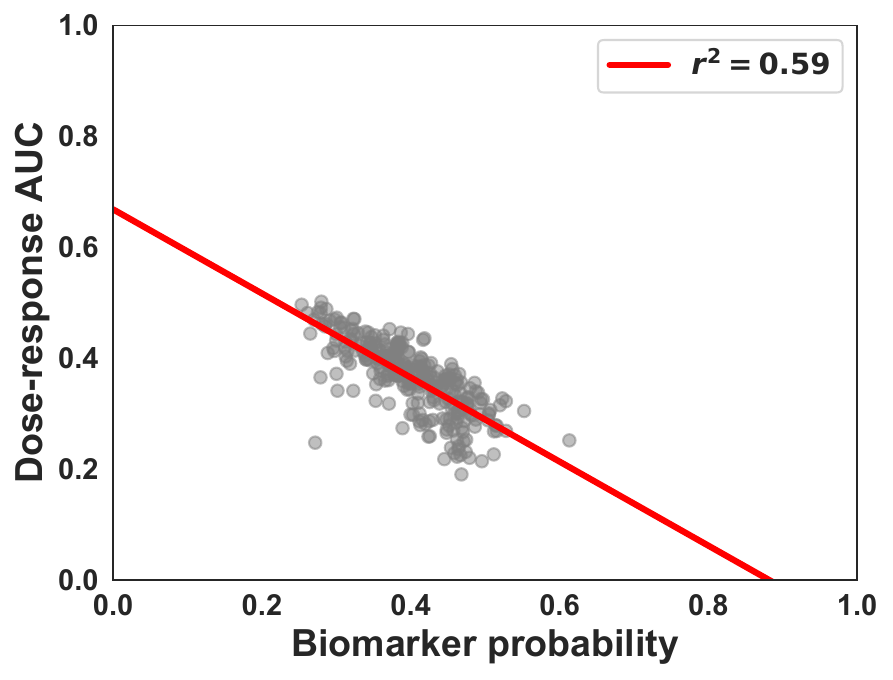}\caption{\label{fig:gene-drug:pten}PTEN CNV + Neratinib}\end{subfigure}
\caption{\label{fig:gene-drug} Top two biomarker results correlating drug sensitivity with biomarker presence. Left: The vIII rearrangement in the EGFR gene is associated with sensitivity to treatment with the drug Trametinib. Right: Copy number variation, typically in the form of a copy number loss, in the tumor suppressor gene PTEN is associated with sensitivity to treatment with the drug Neratinib.}
\end{figure}

\subsection{Uncovering biomarkers associated with drug sensitivity}
\label{subsec:analysis:biomarkers}
To discover potential biomarkers, we correlate the area under the dose-response curve (AUC) with the feature probability. To calculate the AUC, we approximate the curve with a piecewise-linear fit between successive dose intervals,
\begin{equation}
\label{eqn:auc}
\mathrm{AUC} = \frac{1}{T-1}\sum_{t=1}^{T-1} w_i^\top (v_{jt} - v_{j(t+1)}) \, .
\end{equation}
We fit independent linear models to predict the posterior mean $AUC$ values as a function of the biomarker probability. Features are ranked by $r^2$ value and stratified into $sensitivity$ or $resistance$ based on the directionality of their slope. We filter out spurious results by removing features and drugs with standard deviation between samples is less than $0.05$.

\Cref{fig:gene-drug} shows the top two results. Both features are flagged as potential biomarkers of drug sensitivity, indicating that as the probability of the feature increases, the AUC decreases. The top result (\cref{fig:gene-drug:egfr}) associates the vIII rearrangement of the epidermal growth factor receptor (EGFR) gene with sensitivity to treatment with the drug Trametinib. EGFR is involved in many proliferation-inducing signaling pathways, including the important RAS/REF/MEK/ERK pathway \citep{kolch:etal:2015:egfr-pathways}. The vIII rearrangement of EGFR (EGFRvIII) leads to continual activation of EGFR, promoting oncogenesis \citep{guo:etal:2015:egfrviii}. Current approaches to targeting EFGRvIII have not seen major success \citep{an:etal:2018:egfrviii-failures}. The top result flags organoids with the EGFRvIII biomarker as associated with sensitivity to Trametinib, a MEK1/2 inhibitor. This may be due to Trametinib silencing the RAS/REF/MEK/ERK pathway being constitutively activated by the EGFRvIII rearrangement, suggesting a subpopulation of patients with EGFRvIII would benefit from Trametinib.

The second top result (\cref{fig:gene-drug:pten}) is copy number alterations in the phosphatase and tensin homolog (PTEN) gene being associated with sensitivity to treatment with Neratinib. PTEN is a tumor suppressor gene frequently lost in cancer, in particular glioblastoma \citep{koul:2008:pten-glioblastoma} (GBM). The landscape study includes a plurality of GBM organoids and correspondingly the vast majority of the PTEN copy number variations in the dataset are due to PTEN loss. PTEN loss predicts resistance in Trastuzumab \citep{nagata:etal:2004:pten-trastuzumab} and Neratinib has been shown to overcome Trastuzumab resistance in a subset of breast cancer patients \citep{canonici:etal:2013:neratinib-trastuzumab}. The association between PTEN loss and Neratinib sensitivity suggests a possible mechanistic explanation for these effects, as well as a biomarker for other treatment-resistant patients.

\section{Discussion}
\label{sec:discussion}
Multi-sample, multi-drug cancer studies are time and resource intensive. The outcomes from these studies are noisy, often incomplete, observations of biological responses to candidate therapies. Denoising observations and imputing missing experiments is an important step in the scientific analysis and drug discovery pipeline. The Bayesian tensor filtering model we presented in this paper enables scientists to flexibly model dose-response curves with consideration for measurement error and biological constraints on the shape of the curve. While the BTF model is an improvement over the state of the art, we believe there are several improvements that could be made.

The BTF model assumes a regularly-spaced grid. Extensions to irregular grids of dose levels may be possible via a Bayesian group trend filtering extension to an irregular grid approach for scalar trend filtering. The approach taken in \citet{faulkner:minin:2018:bayesian-trend-filtering} was based on integrated Wiener processes. In the case of horseshoe priors, the method requires an approximation \citep{lindgren:rue:2008:second-order-random-walk} that was only tractable up to second-order differences. We have found higher-order priors to not be necessary in our experience, suggesting irregular grid extensions here hold promise. Other methods from the trend filtering literature, such as the falling factorial basis \citep{wang:etal:2014:falling-factorial}, may also be adaptable to the Bayesian group trend filtering case.

The AUC values computed in \cref{subsec:analysis:biomarkers} are likely to be slightly biased due to the piecewise-linear approximation of the true curve. For more precise estimation of the AUC curves, one could adopt the Nadaraya--Watson kernel smoothing of \citet{piegorsch:etal:2012:nonparametric-dose} by introducing pseudo-concentrations for each posterior sample. An alternative approach would be to introduce the pseduo-concentrations as missing data and have the model impute the values directly during posterior sampling with the composite trend filtering enforcing nonstationary smoothness. One could think of the latter approach as the more Bayesian approach whereas the post-hoc smoothing will be computationally faster. The modeling choice is analogous to how we enforce monotonicity directly in the posterior as opposed to post-hoc merging via the PAV approach of \citet{lin:dunson:2014:monotone-gp}.


The current BTF model is computationally intensive. For small scale studies like the pilot study, the model runs in a few hours on a laptop. The landscape study required several days on a compute cluster to perform the hyperparameter search. Relative to the years required for the landscape experiments, the run time is negligible. Nevertheless, offering an alternative inference approach that can scale more efficiently, such as variational inference, may make the BTF model useful for a broader group of scientists.

Finally, BTF does not support adding chemical features about the drugs. In organoid studies, scientists generally know the class of approved and potentially-translatable drugs. Even in high-throughput screening studies for cancer cell lines (i.e. not organoids), only known drugs-- mostly chemotherapy agents-- are typically tested. The goal in these studies is to find biological markers of resistance or sensitivity to the set of available compounds. The molecular feature analysis in \cref{sec:analysis} addresses this. However, if we were trying to discover entirely new drugs, such as might be done at a pharmaceutical company, extending the model to include drug features would be useful. This may be possible by including a second side-information matrix and adding a drug-specific embedding in a manner similar to canonical tensor decomposition.



\begin{small}
\bibliographystyle{plainnat}
\bibliography{arxiv}
\end{small}

\appendix

\section{Local shrinkage updates}
\label{app:inference:horseshoe}
The local shrinkage parameters $\tau_{j\ell}$ can be updated through a double latent variable augmentation trick,
\begin{equation}
\label{eqn:tau_updates}
\begin{aligned}
(\tau_{j\ell} \mid -) &\sim& \mbox{InvGamma}(D+1, \vnorm{\Delta^{(k)}V_j}_2^2 / 2 + 1 / c_{j\ell}) \\
(c_{j\ell} \mid -) &\sim& \mbox{InvGamma}(1, 1/\tau^2_{j\ell} + 1/\phi_{j\ell}) \\
(\phi_{j\ell} \mid -) &\sim& \mbox{InvGamma}(1, 1/c_{j\ell} + 1/\eta_{j\ell}) \\
(\eta_{j\ell} \mid -) &\sim& \mbox{InvGamma}(1, 1/\phi_{j\ell} + 1) \, .
\end{aligned}
\end{equation}
The updates in \cref{eqn:tau_updates} come from the HS+ prior being a two-level horseshoe prior. The inverse-gamma latent variable augmentation for the horseshoe is fast and typically mixes quickly \cite{makalic:schmidt:2015:horseshoe-sampler}.

\section{Gaussian likelihood}
\label{app:inference:gaussian}
When the likelihood is normal, $y_{ijt} \sim \mathcal{N}(w_i^\top v_{jt}, \nu^2)$, where $\nu^2$ is a nuisance parameter, the factor and loading updates are conjugate. Let $\tilde{V} = (v_{1,1}, v_{1,2}, \ldots, v_{1,T}, v_{2,1}, \ldots, v_{M,T})$, and $\Omega^{-1} = \mbox{diag}\{1/\nu^2\}$, then the updates are multivariate normal,
\begin{equation}
\label{eqn:conjugate-updates}
\begin{aligned}
Q^{(i)} &=& (\tilde{V}^\top\Omega^{-1}\tilde{V} + \mbox{diag}(\sigma^{-2}))^{-1} \\
(w_i \mid -) &\sim& \mbox{MVN}\left(Q^{(i)}\tilde{V}^\top\Omega^{-1}\mbox{vec}(Y_i^\top), Q^{(i)}\right) \\
\mathcal{T}^{(j)} &=& \mbox{diag}(1/(\rho^2\tau^2_j)) \\
\Sigma^{(j)} &=& (I_D \otimes \Delta^\top\mathcal{T}\Delta) + (W \otimes I_T)^\top \Omega^{-1} (W \otimes I_T) \\
(\mbox{vec}(V_j) \mid -) &\sim& \mbox{MVN}(\Sigma^{(j)}(W \otimes I_T)\Omega^{-1} \mbox{vec}(Y_{\cdot j}^\top), \Sigma^{(j)}) \, ,
\end{aligned}
\end{equation}
where \texttt{diag} diagonalizes the given vector, \texttt{vec} is the vectorization operator, and $\otimes$ is the Kronecker product. In both the $w_i$ and $V_j$ updates the precision matrices will be sparse, making sampling from the conditionals computationally tractable.

\section{Binomial and related likelihoods via P{\'o}lya--Gamma augmentation}
\label{app:inference:binomial}
When the likelihood is binomial, $y_{ijt} \sim \mbox{Bin}(n_{ijt}, 1/\{1+e^{w_i^\top v_{jt}}\})$, where $n_{ijt}$ is a nuisance parameter, the updates are conditionally conjugate given a P{\'o}lya--Gamma (PG) latent variable sample \cite{polson:scott:windle:2013:polya-gamma},
\begin{equation}
\label{eqn:polya-gamma-updates}
\begin{aligned}
(\psi_{ijt} \mid -) \sim \mbox{PG}(n_{ijt}, w_i^\top v_{jt}) \, , &\qquad (w_i \mid -) \sim N(m_{\psi_i}, \Sigma_{\psi_i}) \, ,
\end{aligned}
\end{equation}
where $\Sigma_{\psi_i} = (\tilde{V}^\top\Psi_i\tilde{V} + \sigma^{-2} I)^{-1}$, $m_{\psi_i} = \Sigma_{\psi_i} \tilde{V}^\top \kappa$, $\Psi_i = \mbox{diag}(\psi_{(i,1,1)}, \ldots, \psi_{(i, M, T)})$, and $\kappa = (y_{(i,1,1)} - n_{(i,1,1)}/2, \ldots, y_{(i,M,T)} - n_{i,M,T}/2)$. The updates for $V$ follow analogously. PG augmentation can be applied to binomial, Bernoulli, negative binomial, and multinomial likelihoods, among others.

\def\Ecal{\mathcal{E}}
\def\X{\mathcal{X}}
\def\L{\mathcal{L}}
\def\N{\mathcal{N}}

\def\unif{\text{unif}}
\def\pr{\text{Pr}}

\newcommand{\ind}{\textbf{1}}

\section{GASS Convergence Proof}
\label{app:GASS:proof}

Recall the GASS sampling algorithm.

\begin{itemize}
	\item Start in some state $x_0$:
	\item For $t = 0, 1, 2, \ldots$:
	\begin{itemize}
		\item Sample $v \sim \N(0, \Sigma)$ and $u \sim \text{unif}([0,1])$.
		\item Compute the sub-ellipse \[ \Ecal_{x_t,v} \ = \ \{ x' = (x_t - \mu) \cos \theta + v \sin \theta + \mu \ : \ \theta \in [0,2\pi] \text{ and } x_t \in S \}.\]
		\item Compute the region $\X_{x_t,u} \ = \ \{ x \in \R^d \ : \ \L(x) \geq \L(x_t) + \log(u) \}$.
		\item Sample the new state $x_{t+1} \sim \unif(\Ecal_{x_t,v} \cap \X_{x_t,u})$.
		\end{itemize}
\end{itemize}
where $\L(\cdot)$ is some log-likelihood function such that $L(x) > -\infty$ for all $x \in \R^d$, $\N(\cdot; \mu, \Sigma)$ is the multivariate normal density, $S \subset \R^d$ is some subset of $\R^d$ with positive Lebesgue measure (in our case it will be an area defined by some linear inequalities).

In this section, we show that the iterates $x_t$ of the GASS chain converge to the correct stationary distribution, whose density is given by
\begin{equation}
\label{eqn:stationary distribution}
P(x) \ = \  \frac{1}{Z} \exp(\L(x)) \N(x; \mu, \Sigma) \ind[x \in S]
\end{equation}
where $Z$ is the normalizing constant to make the density integrate to one.

To begin, we introduce some notation. We say $T(x, \cdot)$ is a Markov transition if $T(x,\cdot)$ is a probability density over $\R^d$ for all $x \in \R^d$. Let $T^n$ denote the transition operator applied $n$ times. Note that $T$, along with a starting state $x_o\in \R^d$, induces a Markov chain $(X_n)_{n=0}^\infty$, where 
\[ \pr(X_{t+n} \in A | X_{t} = x) \ = \ \int_{A} T^n(x, y) \, dy \ =: \ T^n(x,A) \] 
for all measurable $A \subset \R^d$. The \emph{total variation distance between} two probability distribution $\mu, \nu$ is given by
\[ \| \mu(\cdot) - \nu(\cdot) \|_{TV} \ := \ \sup_{A \subseteq \R^d} |\mu(A) - \nu(A)|.  \]
We say a distribution $\pi$ is a \emph{stationary} distribution of $T$ if for all measurable $x, y \in \R^d$, we have 
\[ \int_{x\in \R^d} \pi(dx) T(x, dy) \ = \ \pi(dy). \]
We also say $T$ is \emph{reversible} with respect to a distribution $\pi$ if for $x, y \in \R^d$,
\[ \pi(dx) T(x, dy) \ = \ \pi(dy) T(y, dx).  \]
Let $\phi$ be a non-zero $\sigma$-finite measure, we say that $T$ is \emph{$\phi$-irreducible} if for all subsets $A \subset \R^d$ with $\phi(A) > 0$ and all $x \in \R^d$, there exists a positive integer $n$ such that $T^n(x,A) >0$.

Finally, we say that a Markov chain with transition operator $T$ and stationary distribution $\pi$ is \emph{aperiodic} if there does not exist a partition $A_1, \ldots, A_m$ of $\R^k$ such that $\pi(A_1), \ldots, \pi(A_m) > 0$ and 
\[  T(x, A_{(t \text{ mod } m)+1}) \ = \ 1 \] for all $t = 1, \ldots, m$ and $x \in A_t$.

The following result, which is a simplified version of Theorem 4 in~\cite{RR04}, tells us that reversibility, irreducibility, and aperiodicity are enough to guarantee convergence to stationarity.

\begin{theorem}
\label{thm: convergence conditions}
Suppose $T$ is a transition operator and $\pi$ is a probability distribution over $\R^d$ satisfying
\begin{itemize}
	\item[(i)] $T$ is \emph{reversible} with respect to a distribution $\pi$,
	\item[(ii)] $T$ is \emph{$\pi$-irreducible}, and
	\item[(iii)] $T$ is \emph{aperiodic},
\end{itemize}
then for $\pi$-a.e. $x \in \R^d$, we have that $T^n(x, \cdot)$ converges to $\pi(\cdot)$ in total variation, i.e. 
\[ \lim_{n \rightarrow \infty} \| T^n(x, \cdot) - \pi(\cdot) \|_{TV} \ = \ 0.  \]
\end{theorem}

Thus, to show that the GASS algorithm converges to the correct stationary distribution, it suffices to show that it meets the conditions of Theorem~\ref{thm: convergence conditions}. We start by showing that the GASS transition operator is $P$-irreductible.

\begin{lemma}
\label{lem: irreducibility}
The GASS transition operator is $P$-irreductible, where $P$ is given in equation~\eqref{eqn:stationary distribution}.
\end{lemma}
\begin{proof}
Let $A \subset \R^d$ such that $P(A) > 0$ and let $x \in \R^d$. Since $P(A) > 0$, we know there must exist $x_o \in A$ and $r_o > 0$ such that 
\begin{itemize}
	\item[(i)] $B(x_o, r_o) := \{ z \in \R^d : \|z-x_o\| \leq r_o \} \subset A$ and
	\item[(ii)] $P(B(x_o, r)) > 0$ for each $r \in (0, r_o)$.
\end{itemize}
Furthermore, let $L_{\min} = \inf_{z \in B(x_o, r_o)} \L(z)$. Note that $L_{\min} > -\infty$. Then there is some positive probability that we sample $u \sim \unif([0,1])$ satisfying $\L(x) + \log(u) \leq L_{\min}$. Let us condition on this happening.

From (ii) above, we know that with some positive probability we choose some $v \in B(x_o, r_o/2)$ when sampling from $\N(0, \Sigma)$. Let us condition on this happening. Then there is some subset $B \subseteq [0,2\pi]$ with positive Lebesgue measure such that 
\[ x \cos \theta + v \sin \theta \in B(v, r_o/2) \subseteq A \]
for all $\theta \in B$. Putting it all together, with positive probability, GASS transitions from $x$ to $A$ in a single step.
\end{proof}

We now show that the GASS transition operator is reversible with respect to $P$. Note that the core ideas of this proof already appeared in the reversibility argument of elliptical slice sampling (ESS)~\cite{murray:etal:2010:elliptical-slice-sampling}.

\begin{lemma}
\label{lem: reversibility}
The GASS transition operator is reversible with respect to $P$.
\end{lemma}
\begin{proof}
Let $T(x,\cdot)$ denote the transition operator of GASS. Let $x, x' \in \R^d$. Note that if either $x$ or $x'$ is not in $S$, then we have
\[ P(dx) T(x, dx') = 0 = P(dx') T(x', dx). \]
Thus, assume that $x,x' \in S$.

Now consider any $v \in \R^d$ such that there exists a $\theta \in [0,2\pi]$ satisfying $x' = (x-\mu) \cos \theta + v \sin \theta + \mu$. Then note that if $v' = v \cos \theta - (x-\mu) \sin \theta$, we may rewrite 
\begin{align*}
x &= (x' - \mu) \cos \theta + v' \sin \theta + \mu \\
v &= (x'-\mu) \sin \theta + v' \cos \theta.
\end{align*}
Note that the transformation $(x',v') \rightarrow ((x' - \mu) \cos \theta + v' \sin \theta + \mu,  (x'-\mu) \sin \theta + v' \cos \theta) = (x,v)$ has Jacobian matrix with the following block structure:
\[ J \ = \ 
\begin{pmatrix}
 I_d \cos \theta  &  I_d \sin \theta \\
 I_d \sin \theta & I_d \cos \theta
\end{pmatrix} \]
We can calculate the determinant of this matrix as
\begin{align*}
\det(J) \ &= \ \det(I_d \cos \theta + (I_d \sin \theta)(I_d \cos \theta)^{-1} I_d \sin \theta) \det(I_d \cos \theta) \\
\ &= \ (\cos^2 \theta + \sin^2 \theta)^d \ = \ 1. 
\end{align*} 
Moreover, we have
\begin{align*}
& \N(x ; \mu, \Sigma) \N(v; 0, \Sigma) \\ 
\ &= \ \N((x' - \mu) \cos \theta + v' \sin \theta + \mu; \mu, \Sigma) \N( (x'-\mu) \sin \theta + v' \cos \theta; \mu, \Sigma)  \\
\ &= \ (2\pi)^{-d} \det(\Sigma)^{-1}  \exp \left( - \frac{1}{2} \left( (x' - \mu) \cos \theta + v' \sin \theta \right)^\top \Sigma^{-1} \left( (x' - \mu) \cos \theta + v' \sin \theta \right) \right. \\
&\left.  \hspace{10em}- \frac{1}{2} \left( (x'-\mu) \sin \theta + v' \cos \theta \right)^\top \Sigma^{-1} \left( (x'-\mu) \sin \theta + v' \cos \theta \right) \right) \\
\ &= \ (2\pi)^{-d} \det(\Sigma)^{-1} \exp \left( - \frac{1}{2} (x' - \mu)^\top \Sigma^{-1} (x' - \mu) \cos^2 \theta - \frac{1}{2} (x' - \mu)^\top \Sigma^{-1} (x' - \mu) \sin^2 \theta   \right. \\
&\left.  \hspace{10em} - \frac{1}{2} v'^\top \Sigma^{-1} v' \cos^2 \theta - \frac{1}{2} v'^\top \Sigma^{-1} v' \sin^2 \theta \right) \\
\ &= \  \N(x' ; \mu, \Sigma) \N(v'; 0, \Sigma)
\end{align*}
Putting the above together, we have
\[ \N(x ; \mu, \Sigma) \N(v; 0, \Sigma) \, dx \, dv \ = \ \N(x' ; \mu, \Sigma) \N(v'; 0, \Sigma) \, dx' \, dv'. \]
Moreover, the elliptical regions satisfy $\Ecal_{x, v} = \Ecal_{x',v'}$.

Finally, define $f(z | x, v)$ as the transition probability density function from $x$ conditioned on having chosen $v$ to form the elliptical region $\Ecal_{x, v}$. Then it is not hard to see that 
\[ \exp(\L(x)) f(x' | x, v) \ = \ \exp(\L(x')) f(x | x', v'). \]
Putting everything together, we have the desired result:
\begin{align*}
P(dx) T(x, dx') &= \frac{1}{Z} \exp(\L(x)) \N(x; \mu, \Sigma)  \int_{\R^d} \N(v; 0, \Sigma) f(x' | x, v) \, dx' \, dv \, dx \\
&= \frac{1}{Z} \exp(\L(x')) \N(x'; \mu, \Sigma)  \int_{\R^d} \N(v'; 0, \Sigma) f(x | x', v') \,  dx' \, dv \, dx \\
&= P(dx') T(x', dx). \qedhere
\end{align*}
\end{proof}

Finally, we show that the GASS transition operator is aperiodic.

\begin{lemma}
\label{lem: irreducibility}
The GASS transition operator is aperiodic.
\end{lemma}
\begin{proof}
From Lemma~\ref{lem: reversibility}, we know $P$ is a stationary distribution of $P$. So take $A_1, \ldots, A_m$ to be any partition of $\R^k$ satisfying $P(A_1), \ldots, P(A_m) > 0$. Then from the proof of Lemma~\ref{lem: irreducibility}, we know that for any $x \in A_1$
\[ T(x, A_1) > 0 \]
where $T$ is the GASS transition operator. Thus, for any $x \in A_1$, we have $T(x, A_2) < 1$, which implies that GASS is aperiodic.
\end{proof}

Putting it all together, we have that the GASS chain converges to the desired distribution.

\begin{theorem}
For $P$-a.e. $x \in \R^d$, the GASS chain starting at $x$ converges to $P$ in total variation distance.
\end{theorem}

\end{document}